\documentclass{article}

\usepackage{arxiv}

\usepackage[utf8]{inputenc} 
\usepackage[T1]{fontenc}    
\usepackage{hyperref}       
\usepackage{url}            
\usepackage{booktabs}       
\usepackage{mathptmx} 
\usepackage{amsfonts}       
\usepackage{amsmath} 
\usepackage{amssymb}  
\usepackage{nicefrac}       
\usepackage{microtype}      
\usepackage{cleveref}       
\usepackage{lipsum}         
\usepackage{graphicx}
\usepackage{natbib}
\usepackage{doi}

\usepackage{epsfig} 
\usepackage{mathptmx} 
\usepackage{subfig}
\usepackage{multirow}
\usepackage{tikz}
\usepackage{comment}
\usepackage{color}
\usepackage{footnote}

\title{Convolution kernel adaptation to calibrated fisheye}

\date{}

\author{ \href{https://orcid.org/0000-0003-2674-4844}{\includegraphics[scale=0.06]{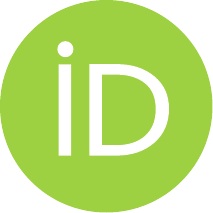}\hspace{1mm}Bruno Berenguel-Baeta*}\thanks{Corresponding author.} \\
	Instituto de Investigacion en Ingenieria de Aragon\\
	Department of Computer Science and Systems Engineering\\
	University of Zaragoza,
	Zaragoza, Spain \\
	\texttt{berenguel@unizar.es} \\
	\And
	Maria Santos-Villafranca*\\
	Instituto de Investigacion en Ingenieria de Aragon\\
	Department of Computer Science and Systems Engineering\\
	University of Zaragoza,
	Zaragoza, Spain \\
	\texttt{m.santos@unizar.es} \\
	\And
	\href{https://orcid.org/0000-0002-8479-1748}{\includegraphics[scale=0.06]{orcid.pdf}\hspace{1mm}Jesus Bermudez-Cameo} \\
	Instituto de Investigacion en Ingenieria de Aragon\\
	Department of Computer Science and Systems Engineering\\
	University of Zaragoza,
	Zaragoza, Spain \\
	\texttt{bermudez@unizar.es} \\
	\And
	\href{https://orcid.org/0000-0002-8949-2632}{\includegraphics[scale=0.06]{orcid.pdf}\hspace{1mm}Alejandro Perez-Yus} \\
	Instituto de Investigacion en Ingenieria de Aragon\\
	Department of Computer Science and Systems Engineering\\
	University of Zaragoza,
	Zaragoza, Spain \\
	\texttt{alperez@unizar.es} \\
	\And
	\href{https://orcid.org/0000-0001-5209-2267}{\includegraphics[scale=0.06]{orcid.pdf}\hspace{1mm}Jose J. Guerrero} \\
	Instituto de Investigacion en Ingenieria de Aragon\\
	Department of Computer Science and Systems Engineering\\
	University of Zaragoza,
	Zaragoza, Spain \\
	\texttt{josechu.guerrero@unizar.es} \\
}
\renewcommand{\shorttitle}{\textit{Fisheye convolutions}}
\newcommand\blfootnote[1]{%
  \begingroup
  \renewcommand\thefootnote{}\footnote{#1}%
  \addtocounter{footnote}{-1}%
  \endgroup
}
\hypersetup{
pdftitle={Convolution kernel adaptation to calibrated fisheye},
pdfsubject={cs.CV},
pdfauthor={B. Berenguel-Baeta, M. Santos-Villafranca, J. Bermudez-Cameo, A. Perez-Yus and J.J. Guerrero},
pdfkeywords={Computer vision; omnidirectional camera; image processing; fish-eye camera; deep learning},
}

\begin{document}
\maketitle
\blfootnote{* Equal contribution}
\blfootnote{A final version of this article can be found at \url{https://proceedings.bmvc2023.org/721/}}

\begin{abstract}
Convolution kernels are the basic structural component of convolutional neural networks (CNNs). In the last years there has been a growing interest in fisheye cameras for many applications. However, the radially symmetric projection model of these cameras produces high distortions that affect the performance of CNNs, especially when the field of view is very large. In this work, we tackle this problem by proposing a method that leverages the calibration of cameras to deform the convolution kernel accordingly and adapt to the distortion. That way, the receptive field of the convolution is similar to standard convolutions in perspective images, allowing us to take advantage of pre-trained networks in large perspective datasets. We show how, with just a brief fine-tuning stage in a small dataset, we improve the performance of the network for the calibrated fisheye with respect to standard convolutions in depth estimation and semantic segmentation. 
The code of the calibrated deformable kernels is publicly available at \small{\url{https://github.com/Sbrunoberenguel/CalibratedConvolutions}}. 

\end{abstract}

\section{Introduction}

\begin{figure}
\centering
	\subfloat[]{\includegraphics[width=0.28\textwidth]{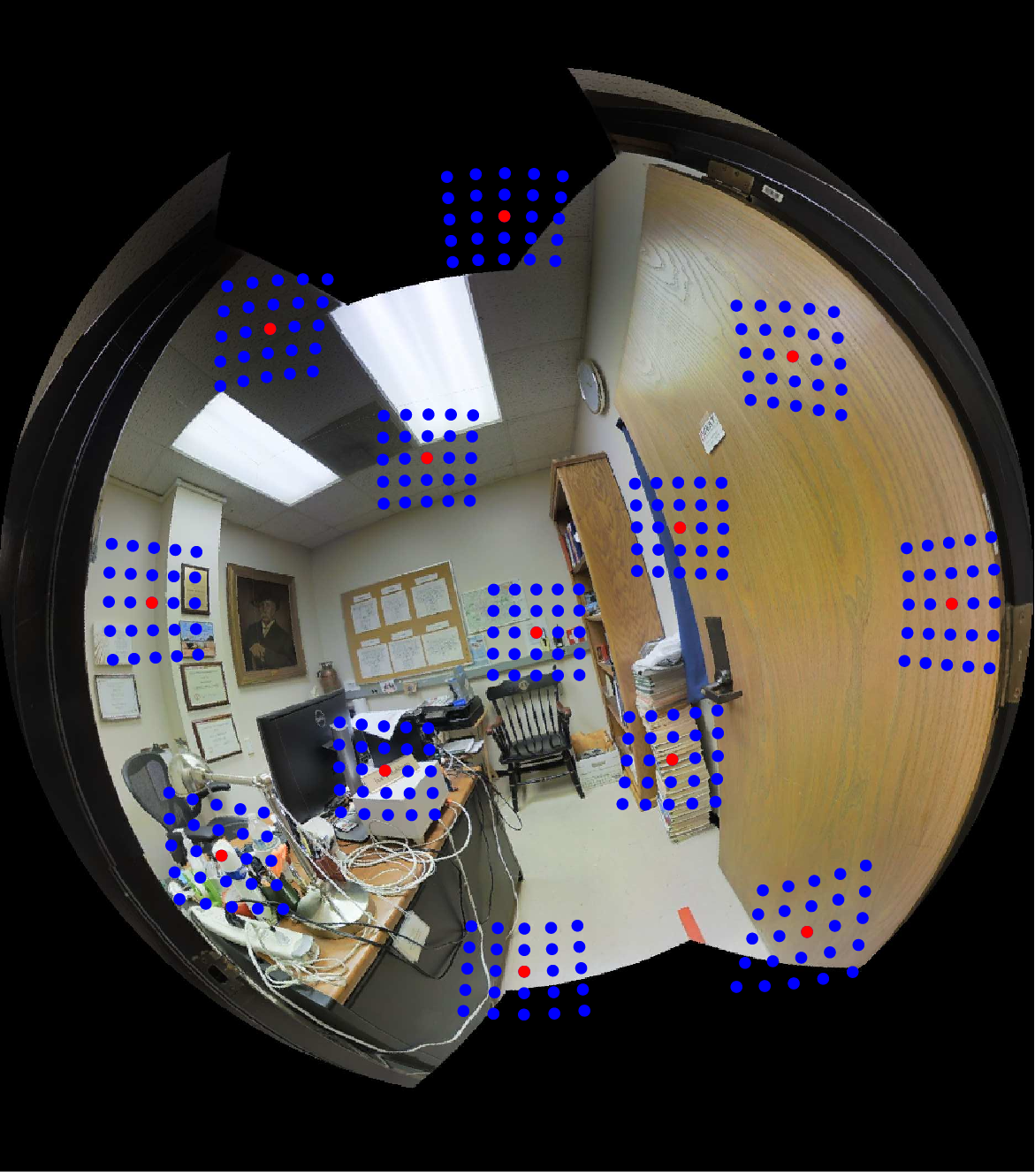}} 
	\subfloat[]{\includegraphics[width=0.35\textwidth]{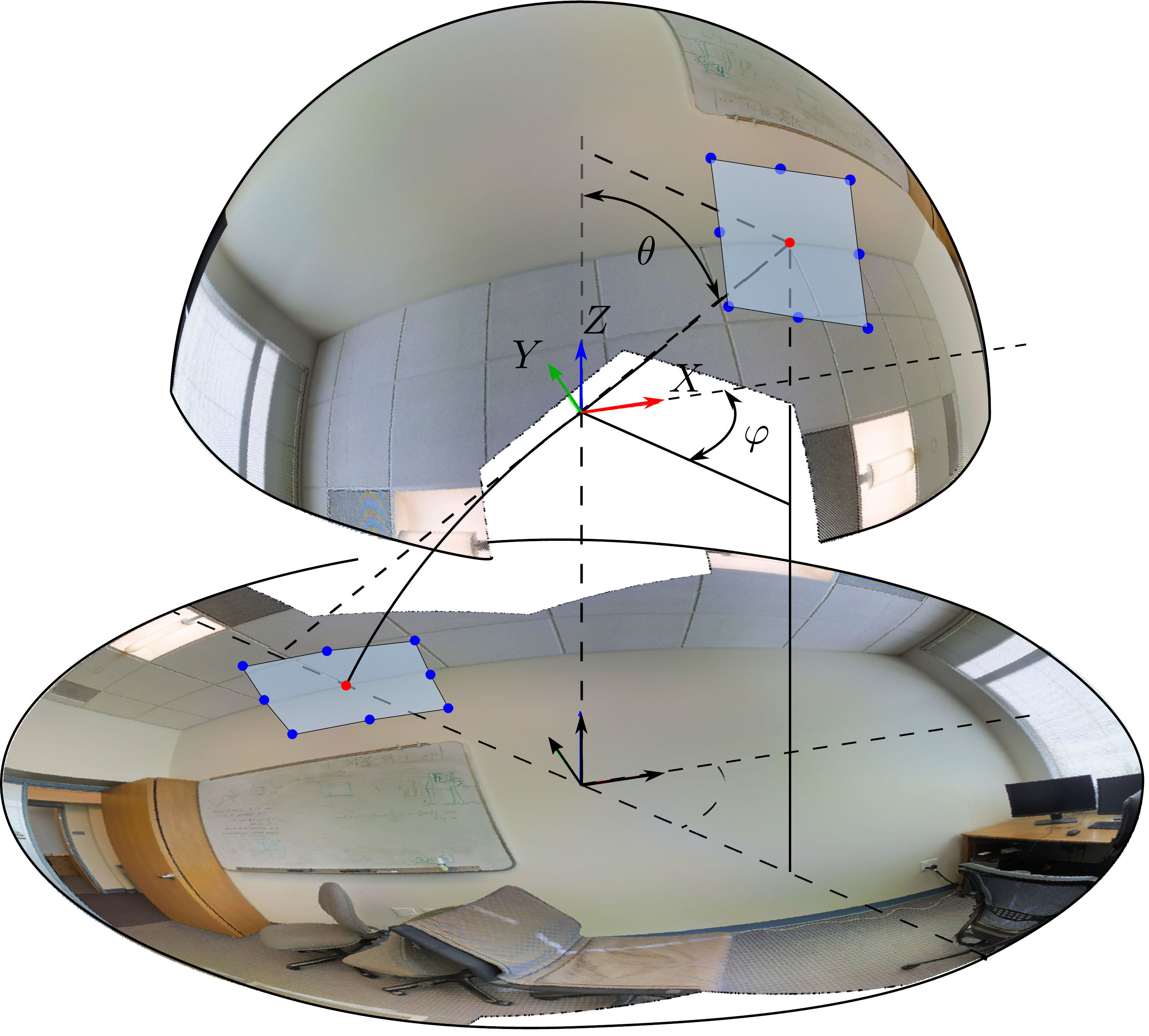}} 
\caption{Overview of how a standard convolution is deformed by the Kannala-Brandt's projection model in a fisheye image. a) Several convolutional kernels adapted to the calibrated fisheye image. b) How the convolutional kernels are computed with the Kannala-Brandt's projection model.}
\label{fig:fisheyemodel}
\end{figure}

Nowadays, Neural Networks are the standard and globally adopted solutions for many machine learning approaches. Among them, Convolutional Neural Networks (CNNs) are the state-of-the-art approaches to handle images and understand what a computer can see. Following classical computer vision algorithms, the first CNNs worked on conventional images (i.e. perspective images). These images provide information of the environment in a relatively small field of view, which is usually enough for the CNNs to achieve great performance in many tasks. This is possible due to the large number of labelled datasets of perspective images, which provide an excellent foundation for these solutions.

On a new trend, unconventional cameras with wider fields of view are becoming popular in many applications and devices such as autonomous vehicles or augmented reality. In particular, we focus on fisheye cameras, which provide several advantages in scene understanding problems. The wide field of view (possibly wider than 180$^\circ$) allows us to achieve a better understanding of the device's surrounding with fewer images. Compared to perspective cameras, with fisheyes it is possible to gather much more spatial information at once, providing more context to the network predictions, especially when the task involve a global understanding of the scene. This is particularly useful in tasks such as semantic segmentation, depth estimation, or room layout estimation.

However, these cameras have some important drawbacks yet to be addressed, which mainly are the reduced number of labelled datasets (mostly because of the difficulty of manually label these images), and the strong distortions induced in large field of view cameras. The last reason causes that existing CNN-based approaches do not transfer well to fisheye images, since the appearance of the elements in the scene is very different to what was learned with perspective datasets. Besides, the contextual information has drastically changed with the much wider field of view, which consequently deteriorates the network's ability to understand the scene. There is definitely a domain gap between perspective and fisheye cameras that needs to be addressed to fully exploit the potential of these devices.

In this work, we propose a method to reduce this domain gap and easily train and use CNNs with highly distorted cameras. We take advantage of the classical perspective CNNs trained with massive datasets, which already provide impressive performance in many different tasks, and adapt these networks to calibrated fisheye cameras. To do so, we propose to substitute the standard convolution operation with deformable convolutions pre-computed with the camera calibration. During the convolution operation, kernels are deformed to accommodate to the distortion depending on the position in the image (see Fig. \ref{fig:fisheyemodel}). With minimal fine-tuning on a small set of data, we can achieve the good performance of well-known CNNs from perspective cameras to fisheye cameras, not needing to create new large datasets specific to each desired camera calibration in order to help the network to learn the distortions. 
The main contributions of this work are:
\begin{itemize}
	\item We present a novel implementation of calibrated deformable convolution for fisheye cameras under the Kannala-Brandt projection model, which could be used with any fisheye camera (even with a field of view wider than 180$^\circ$).
	\item We propose a set of experiments of domain adaptation of well known CNNs for several tasks with fisheye cameras. Particularly, we show results with different fields of view, showing that our method allows great flexibility in the camera configurations with minimal effort.
\end{itemize}

\section{Related work}

In recent years, there has been a surge of using fisheye and 360$^\circ$ cameras since they introduce more information within a single image, which is advantageous when tackling different computer vision tasks such as: scene understanding \cite{gao2022review}, depth estimation \cite{kumar2018monocular} \cite{berenguel2022fredsnet} \cite{li2022omnifusion}, semantic segmentation \cite{berenguel2022fredsnet} \cite{deng2017cnn}, object \cite{guerrero2020s} and pedestrian detection \cite{haggui2021human}, or autonomous driving \cite{toromanoff2018end}, among others \cite{baek2018real} \cite{ni2019vanishing} \cite{perez2019scaled}. There is also an increasing presence of wide-angle cameras such as fisheyes in mobile devices such as phones, or VR headsets and stereo cameras, due to the improved robustness in localization and mapping from the larger field of view.

However, when it comes to deep learning methods applied to wide angle images, one of the most common operations, the convolution, is flawed. The space-varying distortion caused by the image projection models for omnidirectional and wide-angle cameras makes the translational weight sharing ineffective \cite{cohen2018spherical}. Objects appear differently distorted depending on where they are in the image, which makes more difficult for the network to learn each plausible configuration, especially considering different camera calibrations. Therefore, there is still an open challenge about how to use and train traditional CNN architectures with these kinds of images, also considering the lack of large datasets compared to perspective images. 
Some researchers have focused on adapting CNNs to the spherical domain. For instance, Cohen et al. \cite{cohen2018spherical} proposed Spherical CNNs, studying convolutions on the sphere using spectral analysis. Jian et al. \cite{jiang2019spherical} replace conventional convolution kernels with linear combinations of differential operators weighted by learnable parameters on unstructured grids.
Su and Grauman  \cite{su2017learning} aims to adapt a CNN trained on perspective images to the equirectangular domain adjusting the sizes of the rectangular kernels depending on the elevation angle. UniFuse \cite{jiang2021unifuse} proposed to fuse features from Cubemap projection with regular convolutions on equirectangular images on the decoding stage for depth estimation. In an attempt to make more efficient the computation, \cite{zioulis2019spherical} use spherical attention masks to make the model aware of its spherical nature.

On the other hand, different approaches have been proposed to enable CNNs to be more dynamic, improving the performance for specific tasks, as well as extending their applicability to new domains \cite{jeon2017active, dai2017deformable, zhuang2022acdnet}.
For example, \cite{jeon2017active} and \cite{dai2017deformable} focus on convolution units with no fixed shape and learned offsets for the convolution kernel. Jeon and Kim \cite{jeon2017active} use these convolutions to obtain better receptive field for object classification, whereas \cite{dai2017deformable} incorporates spatial deformations into the convolutional operation to be able to handle objects with significant variations in shape or appearance. 
The deformable convolutions \cite{dai2017deformable} were used in \cite{fernandez2020corners} for layout estimation, with not learned but fixed offsets, pre-computed to account for the image distortion that occurs in equirectangular projections. Thus, the receptive field of the convolution filter is undistorted.
The idea of making the convolution “on the sphere” and project the convolution kernel with the equirectangular projection model was also proposed by \cite{tateno2018distortion} with their \textit{distortion-aware convolutions}, introducing a pipeline of transfer learning from learned convolution filters in perspective images applied for depth estimation in equirectangular panoramic images. Strategies like transfer learning or domain adaptation have been successfully used in the past \cite{donahue2014decaf,  sharif2014cnn}, and we believe it could alleviate the absence of datasets for our task. This approach was recently explored for distortion-aware convolutions in \cite{artizzu2023omni}.

However, most works disregard the specific calibration parameters, mostly because they use very simple image projections such as equirectangular projection, that directly map azimuth and elevation angles in pixel locations. Thus, the research on these distortion-aware convolutions for other configurations such as fisheyes is scarce and very specific to the camera pose \cite{meng2021distortion}. Only a few works directly deal with the camera calibration parameters into the convolutions, like CAMConvs \cite{facil2019cam}. In this work, we aim to breach that gap, introducing novel convolutions for radial-distortion models, like the Kannala-Brandt’s projection model \cite{kannala2006generic}, that adapts to the specific calibration of the camera and is apt for any fisheye camera. Drawing inspiration from \cite{fernandez2020corners}, our approach uses deformable convolutions \cite{dai2017deformable} to adapt the convolution filters to the distortion caused by the projection. Up to our knowledge, this is the first work that explicitly deals with calibrated convolutions for radially distorted cameras. We show how, with just a little fine-tuning in a relatively small dataset, classical CNN methods can be adapted to any calibrated camera. Our approach with calibration specific kernel offsets could also be extended to other calibration models and account for their distortion.

\section{Fisheye convolution}

Convolutions are the keystone of CNNs and current computer vision algorithms. The work \cite{dai2017deformable} presents a learned deformable kernel to improve the performance of several CNNs. In this work, we propose to use calibration-based deformable kernels. We use the camera calibration of fisheye cameras to compute the offsets of the kernel positions and adapt the convolution to the distortion of these images. In this section we summarize the projection model used and how we apply the distortion to the kernels.

\subsection{Fisheye projection model} 

In the literature of non-conventional cameras we find many works that propose mathematical models of several projections. Considering fisheye cameras, we can also find several models such as the \textit{equidistance}, \textit{stereographic}, \textit{orthogonal} or \textit{equisolid angle}. Each of these models propose a different non-linear function to fit the distortion of the projecting rays of different lens configurations and geometries. In our case, we aim to cover a wider set of projection models, so we use empirical models, which are more flexible to different or unknown projection models.

In the field of empirical models for camera calibration, two rise above the others: Scaramuzza's \cite{scaramuzza2006} and Kannala-Brandt's \cite{kannala2006generic}. Both models propose a high-order polynomial function to model the camera distortion. In this work, we use the second one, Kannala-Brandt's \cite{kannala2006generic}, since it is the most extended in the calibration of commercial devices.

The full Kannala-Brandt model is explained in \cite{kannala2006generic}, where they present a radially symmetric model and a full model were the radial asymmetry of lenses is taken into account. For this work we use the first one, assuming that the radial asymmetry error is much lower than the pixel precision that can be obtained from an image. So, assuming a radially symmetric model, \cite{kannala2006generic} define the forward projection model as:

\begin{equation} \label{eq:KBForward}
\begin{pmatrix}
u\\ v
\end{pmatrix} =
d(\theta) 
\begin{pmatrix}
f_x \cos \varphi \\ f_y \sin \varphi
\end{pmatrix} +
\begin{pmatrix}
c_x \\ c_y
\end{pmatrix}
\end{equation}\noindent
where $(u,v)$ are the pixel coordinates in the image, $(c_x,c_y)$ are the pixel coordinates of the optical center, $(f_x,f_y)$ is the focal length on each axis, $(\theta,\varphi)$ are the spherical coordinates (see Fig. \ref{fig:fisheyemodel} (b)) of the incoming ray, $d(\theta) = k_1 \theta + k_2 \theta^3 + k_3 \theta^5 + k_4 \theta^9$ is the high order polynomial function and $[k_1,k_2,k_3,k_4]$ the Kannala-Brandt calibration parameters.

The back projection model is computed as:
\begin{equation} \label{eq:KBBackward}
\varphi = \arctan \frac{m_y}{m_x} ; \;\;\;
\theta = d^{-1}\left(\sqrt{m_x^2+m_y^2}\right),
\end{equation}
where $m_x = (u-c_x)/f_x$, $m_y = (v-c_y)/f_y$ and $d^{-1}$ is computed iteratively.

\subsection{Fisheye calibrated kernel}

Deformable convolutions were presented in \cite{dai2017deformable}, where the authors propose a method to learn offsets in the kernels for a better adaptation of the CNN to the task at hand. On the other hand, using the calibration of perspective cameras on CNNs has also been addressed by \cite{facil2019cam}, obtaining improvements in the performance of the network. In this section, we present our implementation of a camera-calibrated kernel for non-linear projection models, adapting the kernel to the fisheye distortion of the Kannala-Brandt projection model.

Let $\mathcal{K}$ be a $(k_i \times k_j)$ rectangular kernel where $(k_i,k_j) \geq 1$ are an odd number and $(u_0, v_0)$ is the anchor pixel around which we will apply the convolution kernel. We define the coordinates of each element of $\mathcal{K}$ as:
$ \hat{p}_{ij} = \left(i, j, d \right)^T $
where $i$ is in range $\left[ -\frac{k_{i}-1}{2},\frac{k_{i}-1}{2} \right]$; $j$ is in range $\left[ -\frac{k_{j}-1}{2},\frac{k_{j}-1}{2} \right]$; and $d$ is the focal distance of $\mathcal{K}$. We assume that the standard kernel has the same behaviour as in a perspective camera, so we compute the focal distance as a function of the field of view of the kernel, $\alpha$, which we linearly map from the fisheye camera field of view, $\Phi$, as:
$	d = \frac{k_i}{2 \tan \frac{\alpha}{2}} $
where $\alpha = \frac{k_i}{W} \Phi$ where $W$ is the size of the feature map.

We project each kernel point into the unit sphere surface by normalizing the vectors. Then, we want to go back to the pixel domain using the forward projection model of the camera, Eq. \ref{eq:KBForward}. However, the resolution of the feature maps of a network (almost) always differs from the input image of the network, which is the resolution of the calibration parameters of the camera. Besides, we have to align the anchor of the kernel with the pixel we want to apply the kernel.

To solve the multi-scale/multi-resolution problem, we compute an scaling factor for each resolution that relates the calibration resolutions with the current feature map. The scaling factor is defined as $s = \frac{W_c + p_w}{W_{FM}}$, where $W_c$ is the camera resolution width, $W_{FM}$ is the feature map width and $p_w$ is the width of an additional padding to the input image. We use this padding to set a fixed input resolution to our network in case of different image resolutions. With the scaling factor, we re-compute the calibration parameters for the kernel as:
\begin{equation}
	\begin{pmatrix}
	cx_k \\ cy_k
	\end{pmatrix} = 
	\begin{pmatrix}
	c_x \; \frac{W_{FM} - p_w/s}{W_c} \\
	c_y \; \frac{H_{FM} - p_h/s}{H_c} \\
	\end{pmatrix}; 
	\;\;
	\begin{pmatrix}
	fx_k \\ fy_k
	\end{pmatrix} = 
	\begin{pmatrix}
	f_x \; \frac{W_{FM} - p_w/s}{W_c} \\
	f_y \; \frac{H_{FM} - p_h/s}{H_c} \\
	\end{pmatrix},
\end{equation}\noindent
where $(cx_k,cy_k), (fx_k,fy_k), (W_{FM}, H_{FM})$ are the coordinates of the optical center, focal lengths and resolution of the feature map and $(W_c,H_c)$ the resolution of the fisheye camera.

Once the calibration parameters are adjusted to the feature map resolution, we compute the projecting ray of the anchor pixel $(u_0,v_0)$ using Eq.\ref{eq:KBBackward} and rotate the projecting rays of $\mathcal{K}$ to meet the orientation. With the $\mathcal{K}$ in the correct position, we project again each element of the kernel into the fisheye plane with Eq.\ref{eq:KBForward}, obtaining the new locations of the kernel in the fisheye image (or feature map).
From this implementation, we obtain a convolution kernel that adapts its shape with the distortion of the camera following the radially symmetric projection model (see Fig. \ref{fig:fisheyemodel}).

\section{Experiments}
For the experimental part, we evaluate the calibrated convolutions against the standard convolutions on fisheye cameras. For that purpose, we use a well known CNN, U-Net \cite{ronneberger2015u}, on two different tasks to evaluate the performance of the proposed kernels. We want to avoid current architectures where convolutions are mixed with other components as recurrent blocks \cite{hochreiter1997long} or attention mechanisms \cite{vaswani2017attention} to evaluate only the impact of the convolutions in the overall performance. 
For a fair comparison between convolutions, we propose the following set-up for the experiments: 
    1) We train the CNN, with standard convolutions, on perspective images of the Stanford dataset \cite{armeni2017joint} and use these weights as baseline; 
    2) We evaluate the network with standard convolutions and calibrated convolutions on fisheye images obtained from the Stanford dataset \cite{armeni2017joint}; 
    3) With the baseline as pretrained weights, we fine tune the network with fisheye images in two different situations: one fine tune is made with standard convolutions and other with the proposed calibrated convolutions;
    4) After fine tuning, we evaluate again both networks on the fisheye images.
To extend our comparison, we also fine tune and evaluate the network with standard convolutions on rectified images (from fisheye to perspective), an alternative approach only applicable when field of view is <180º. More details of the rectification process and full experiment is available in the supplementary material.

Training and fine tuning is made in the Stanford dataset following the $\#1$ folder split, taking the \textit{Area 5} only for evaluation and the others as training and validation sets. 
Perspective images are taken from the original dataset, where we find around 70k images with depth and semantic information. For simplicity, we define this dataset as \textit{Pers} in the experiments. 
Fisheye images are randomly synthesized from the panoramic dataset in different orientations and with known calibration. We have generated 11.2k images for two different fisheye calibrations (i.e. 5.6k images of each calibration). For simplicity, we define these datasets by the field of view of the camera, such as: \textit{F165} defines the dataset of fisheye images with a field of view of $165^\circ$ and \textit{F195} with a field of view of $195^\circ$. 

\subsection{Monocular depth estimation}

\begin{table}[t]
\centering
\footnotesize	
	\begin{tabular}{c|cc|ccccccc}
 	 & \multicolumn{1}{c|}{Dataset} & \multicolumn{1}{c|}{Kernel} & 
            MRE $\downarrow$     & MAE $\downarrow$  & RMSE $\downarrow$     & RMSE$_{log}$ $\downarrow$   & 
            $\delta^1$ $\uparrow$   & $\delta^2$ $\uparrow$   & $\delta^3$ $\uparrow$ \\ \hline
	\multicolumn{1}{c|}{\multirow{6}{*}{BL}}  & \multicolumn{1}{c|}{\textit{Pers}} & Standard	& 
            0.1538 & 0.2462 & 0.4908 & 0.1146 & 0.6391 & 0.8678 & 0.9407 \\ \cline{2-10}
	       & \multicolumn{1}{c|}{\multirow{2}{*}{\textit{F195}}} & Standard	& 
            0.2726 & 0.3914 & 0.4943 & 0.2408 & 0.4186 & 0.7027 & 0.8517 \\
		\multicolumn{1}{c|}{} & \multicolumn{1}{c|}{} & Calibrated 	& 
            0.2922 & 0.4133 & 0.5262 & 0.2604 & 0.3839 & 0.6745 & 0.8371 \\\cline{2-10}
		  & \multicolumn{1}{c|}{\multirow{3}{*}{\textit{F165}}} & Standard	& 
            0.2670 & 0.3790 & 0.5005 & 0.2324 & 0.4377 & 0.7198 & 0.8579 \\
		\multicolumn{1}{c|}{} & \multicolumn{1}{c|}{} & Calibrated 	& 
            0.2798 & 0.3971	& 0.5216 & 0.2441 & 0.4019 & 0.6998 & 0.8500 \\
		\multicolumn{1}{c|}{} & \multicolumn{1}{c|}{} & Rectified 	& 
            0.8595 & 0.6412	& 0.9714 & 0.4178 & 0.3000 & 0.5509 & 0.7305 \\
            \hline \hline
	\multicolumn{1}{c|}{\multirow{5}{*}{FT}}  & \multicolumn{1}{c|}{\multirow{2}{*}{\textit{F195}}} & Standard	& 
            0.2432 & 0.3729	& 0.4023 & 0.2022 & 0.4241 & 0.7277 & 0.8827 \\
		\multicolumn{1}{c|}{} & \multicolumn{1}{c|}{} & Calibrated 	& 
            \textbf{0.2017} & \textbf{0.3159}	& \textbf{0.3418} & \textbf{0.1575} & \textbf{0.5450} & \textbf{0.7972} & \textbf{0.9075} \\ \cline{2-10}
		  & \multicolumn{1}{c|}{\multirow{3}{*}{\textit{F165}}} & Standard	& 
            0.2508 & 0.3582	& 0.4040 & 0.1899 & 0.4962 & 0.7628 & 0.8879 \\
		\multicolumn{1}{c|}{} & \multicolumn{1}{c|}{} & Calibrated 	& 
            \textbf{0.2505} & \textbf{0.3561}	& \textbf{0.3875} & \textbf{0.1865} & \textbf{0.4992} & \textbf{0.7648} & \textbf{0.8884} \\ 
            \multicolumn{1}{c|}{} & \multicolumn{1}{c|}{} & Rectified	& 
            0.7758 & 0.5999	& 0.8933 & 0.3661 & 0.3016 & 0.5710 & 0.7618 \\
            \hline
\end{tabular}
\caption{Monocular depth estimation with standard and calibrated convolutions for U-Net neural network. BL: Base Line; FT: Fine Tuned. Best metric for each fisheye calibration is in bold.}
\label{tab:monoDepth:UNET}
\end{table}

The first task we evaluate is monocular depth estimation from single images. 
The convolutional network has been trained for 50 epochs on the Stanford dataset with perspective images, which has taken around 100 hours to complete. This network is our baseline (BL) for the experiment. Then, the fine tuning (FT) has been made on top of this training for less than 10\% of the training time. The network has been fine tuned for 20 epochs in the fisheye dataset, taking between 2-4 hours. Fine-tuning time changes with the resolution of the fisheye images, being different for each field of view. 

\begin{figure}[t]
    \centering
    \subfloat[F165 RMSE]{\includegraphics[trim={0.9cm 0.9cm 0.9cm 0.9cm},width=0.24\textwidth]{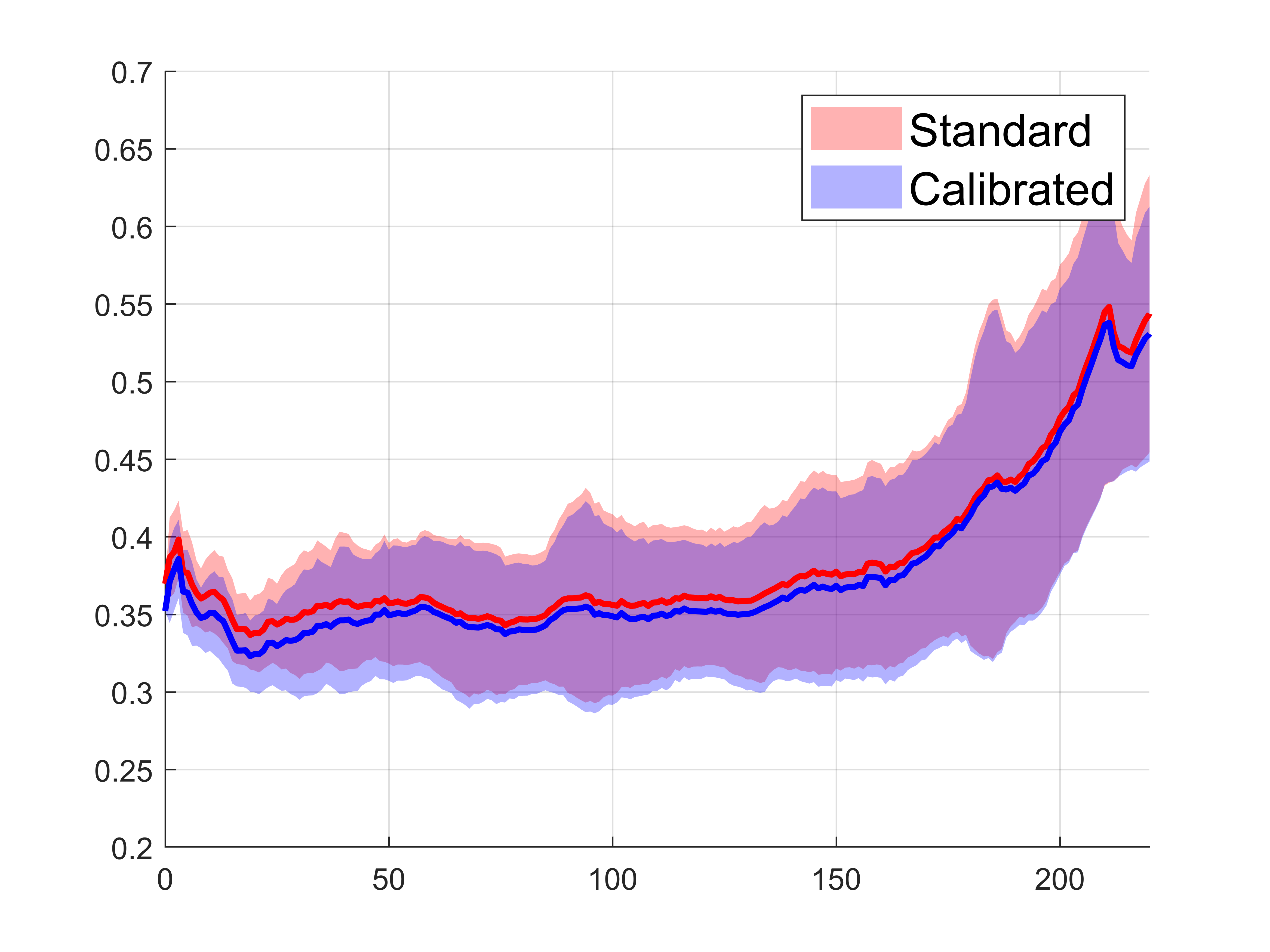}}
    \subfloat[F165 $\delta^1$]{\includegraphics[trim={0.9cm 0.9cm 0.9cm 0.9cm},width=0.24\textwidth]{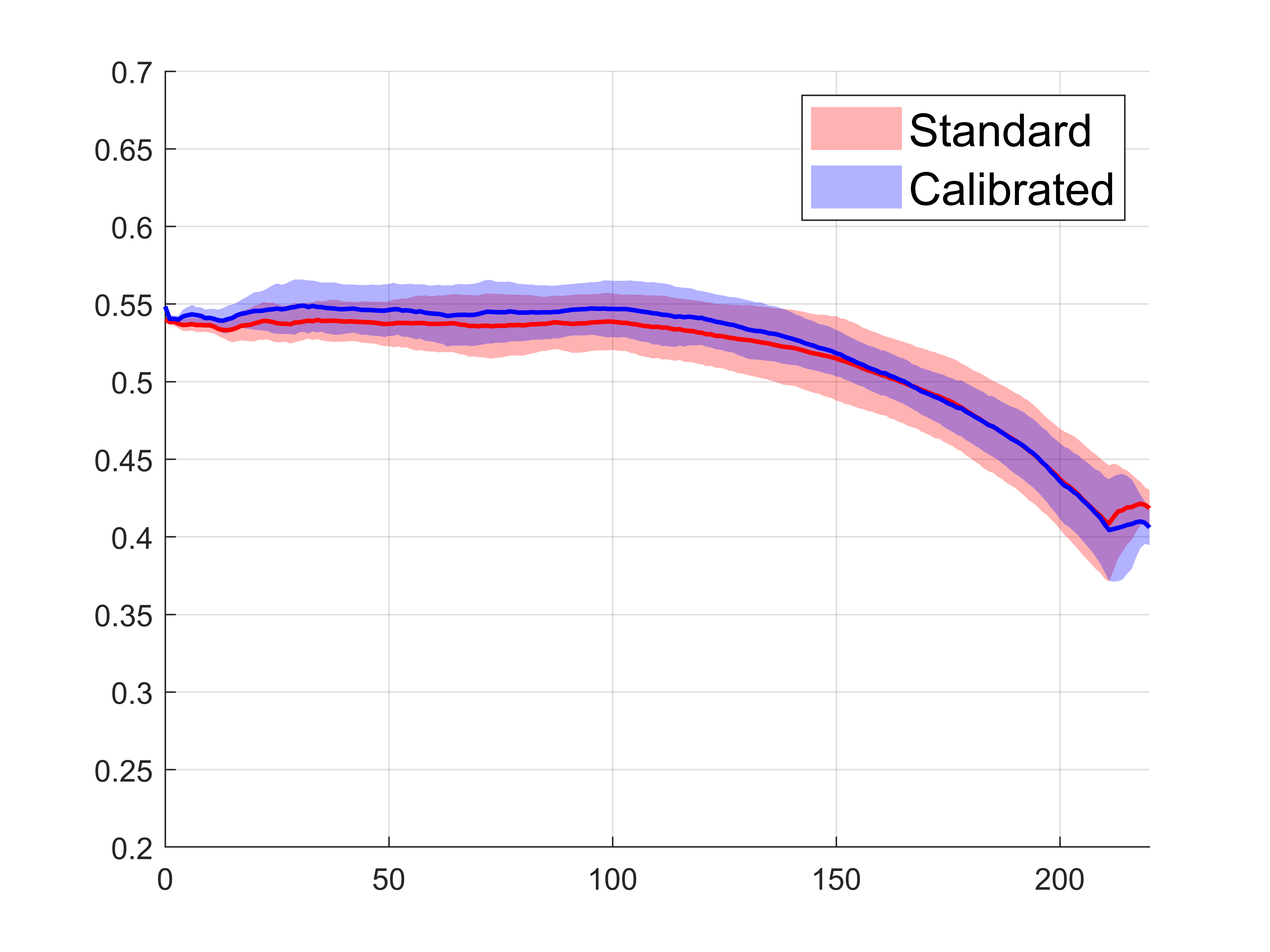}}
    \subfloat[F195 RMSE]{\includegraphics[trim={0.9cm 0.9cm 0.9cm 0.9cm},width=0.24\textwidth]{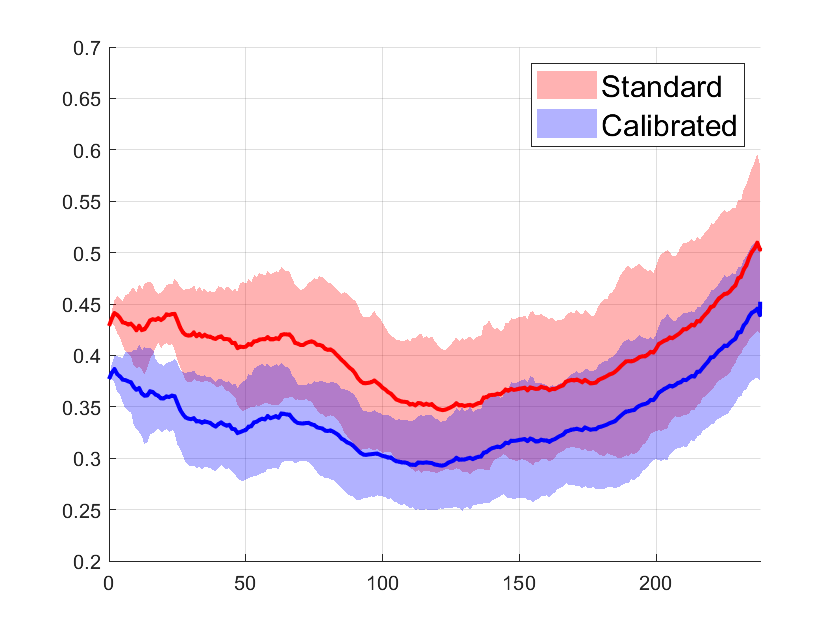}}
    \subfloat[F195 $\delta^1$]{\includegraphics[trim={0.9cm 0.9cm 0.9cm 0.9cm},width=0.24\textwidth]{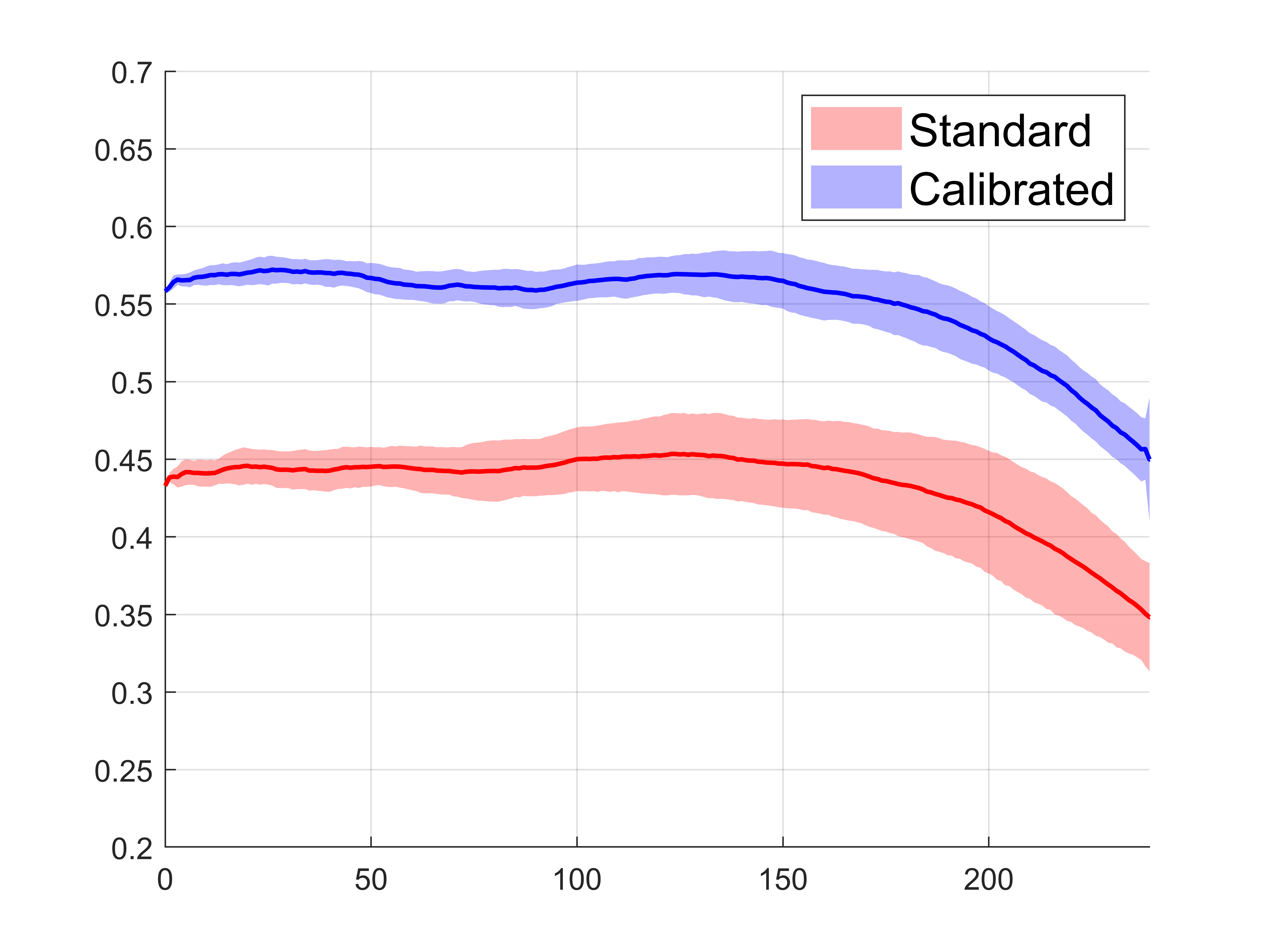}}
    \caption{Comparison and results of depth estimation with U-Net neural network with standard (red) and calibrated (blue) convolutions. The x-axis defines the distance of the pixels to the optical center and the y-axis the computed error, defined as mean and one standard deviation.}
    \label{fig:depth:UNET}
\end{figure}

\begin{figure}[t]
    \centering
    \subfloat[F165 RGB]{\includegraphics[height=0.24\textwidth]{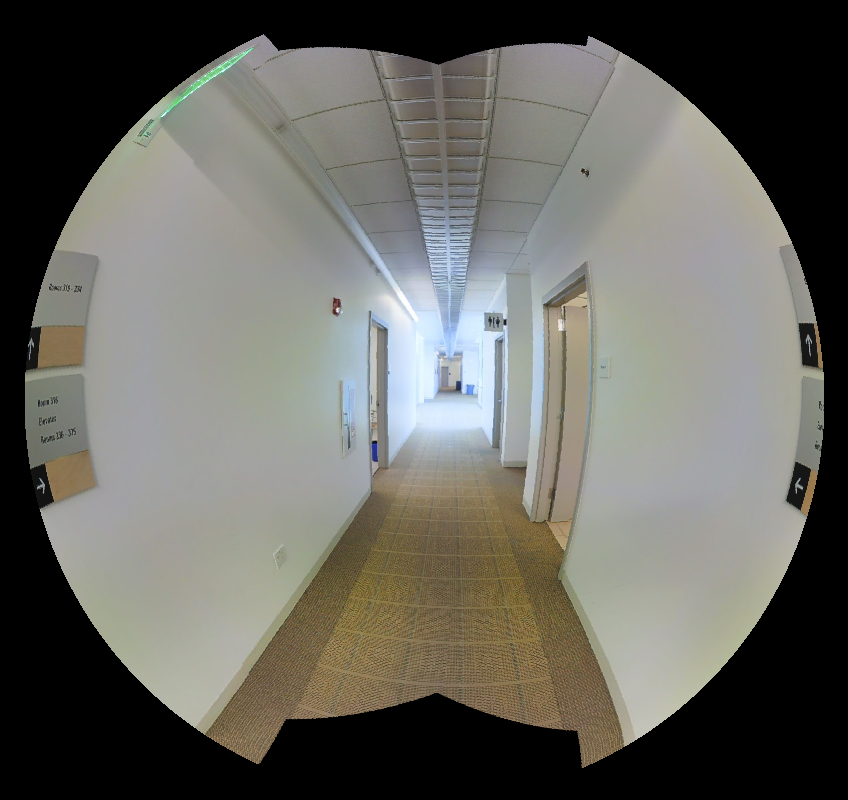}}
    \subfloat[F165 Standard]{\includegraphics[height=0.24\textwidth]{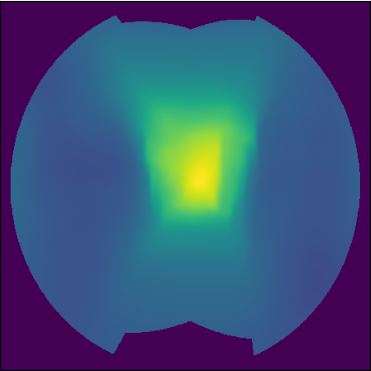}}
    \subfloat[F165 Calibrated]{\includegraphics[height=0.24\textwidth]{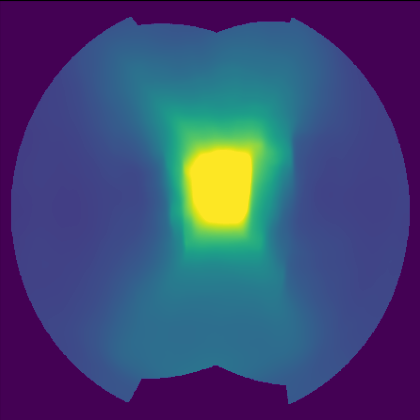}}   		\subfloat[F165 GT]{\includegraphics[height=0.24\textwidth]{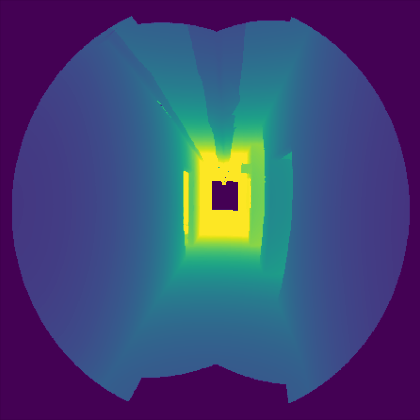}}\\ 
    \subfloat[F195 RGB]{\includegraphics[height=0.223\textwidth]{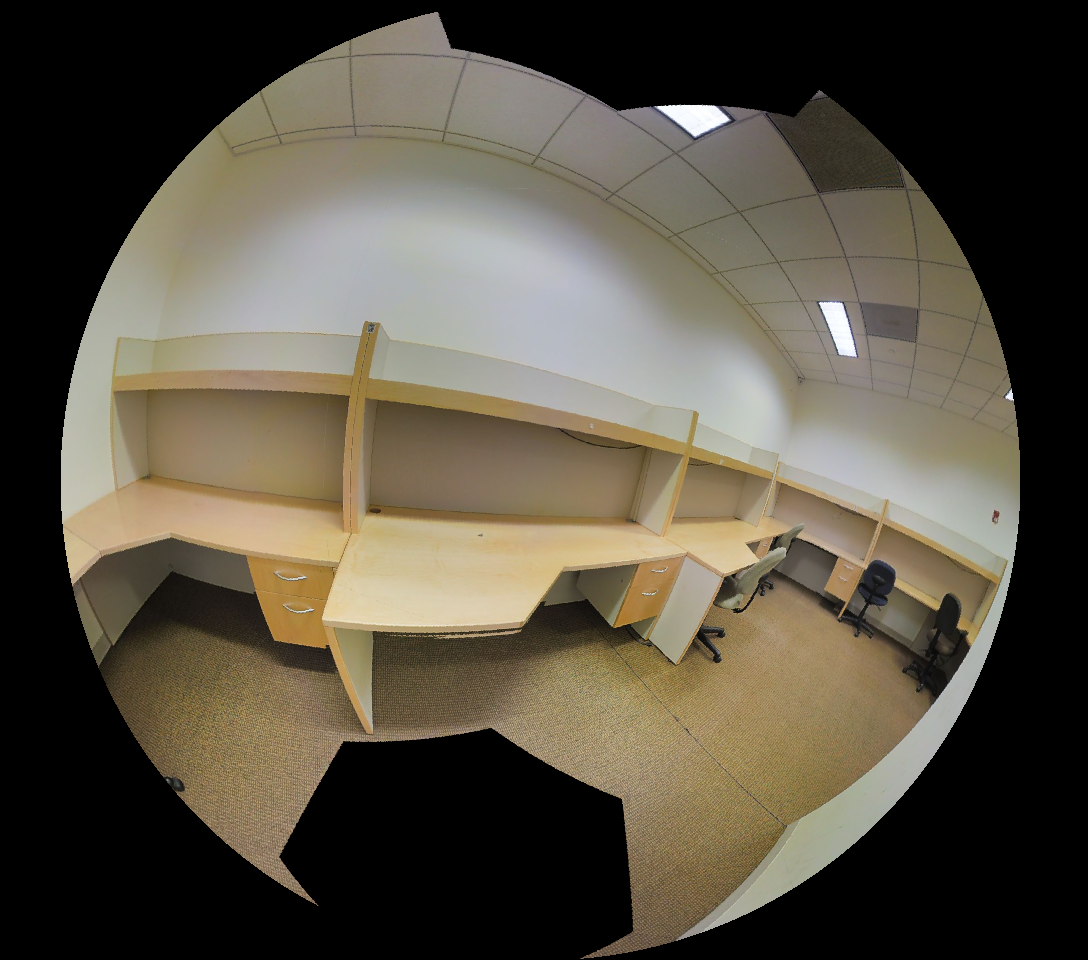}}
    \subfloat[F195 Standard]{\includegraphics[height=0.223\textwidth]{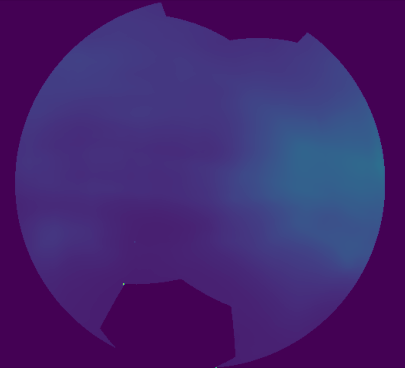}}
    \subfloat[F195 Calibrated]{\includegraphics[height=0.223\textwidth]{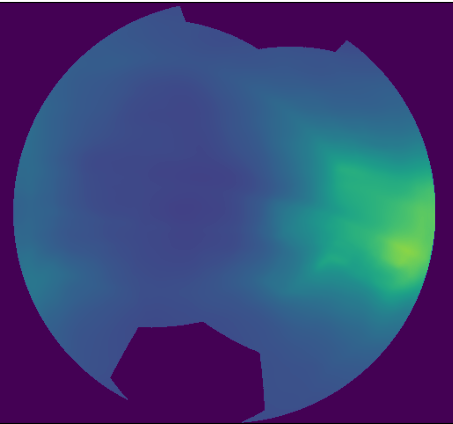}}
    \subfloat[F195 GT]{\includegraphics[height=0.223\textwidth]{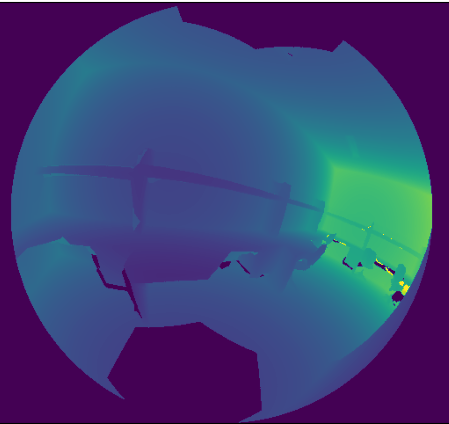}}
    \caption{Qualitative results of monocular depth estimation on different fisheye calibrations. Distance is in a color scale, from colder colors (closer distances) to warmer colors (farther distances).}
    \label{fig:depth:qualy}
\end{figure}

\begin{figure}
    \centering
    \subfloat[Standard Conv]{\includegraphics[width=0.25\textwidth]{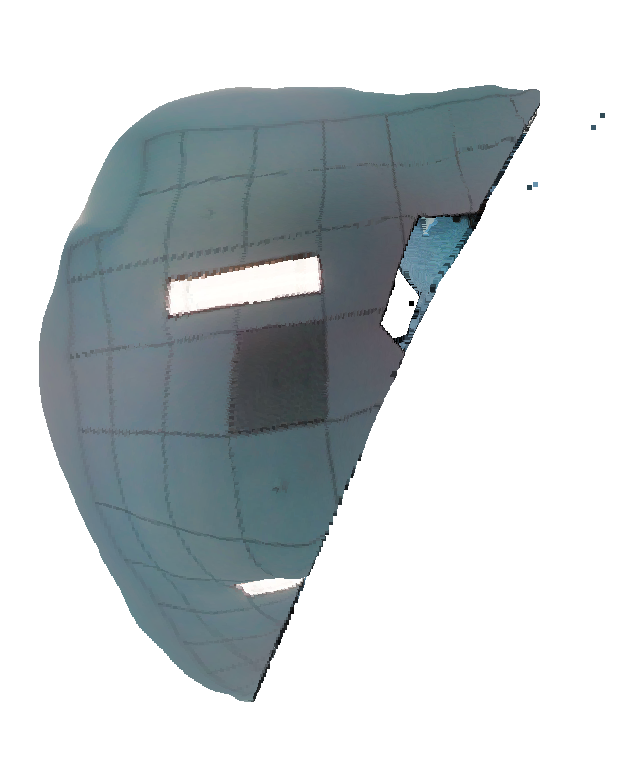}}
    \subfloat[Calibrated Conv]{\includegraphics[width=0.25\textwidth]{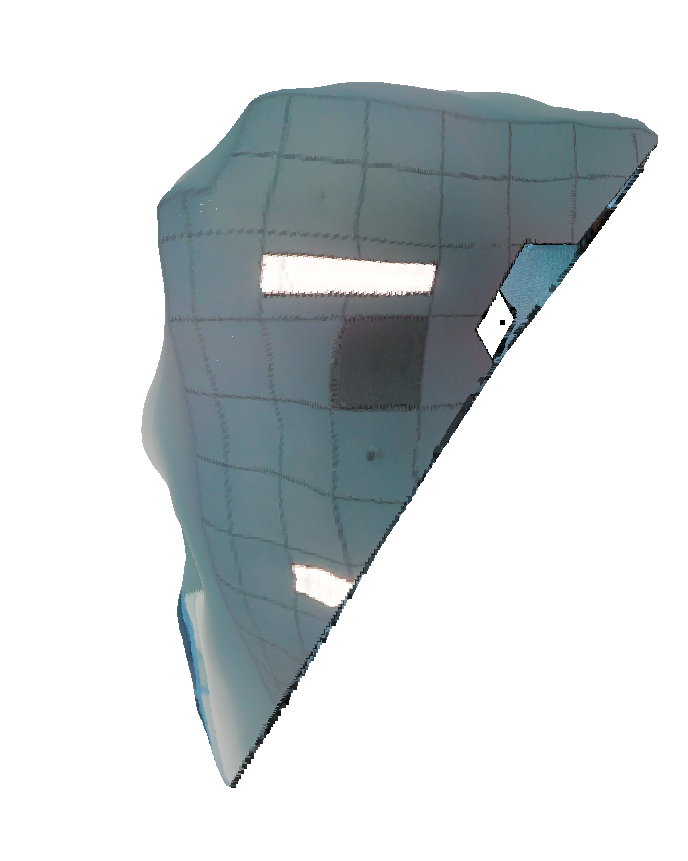}}\hfil
    \subfloat[Ground Truth]{\includegraphics[width=0.25\textwidth]{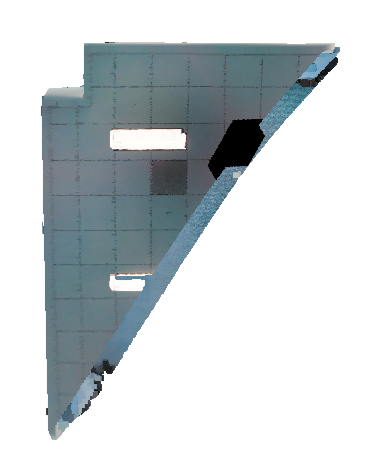}}   
    \caption{Qualitative results of depth estimation for FOV of 195º, top view of a 3D point cloud generated from depth data.}
    \label{fig:depth:pc}
\end{figure}

Results of this experiment are shown in Tab. \ref{tab:monoDepth:UNET}. We use the standard metrics for depth estimation presented in \cite{zioulis2018omnidepth}. 
We also compute the metrics with respect the distance of each pixel to the principal point of the camera (i.e. $d(\theta)$ from equation \ref{eq:KBForward}) to observe the behaviour of each convolution with the increasing distortion of the image. These results are presented in Fig. \ref{fig:depth:UNET} for both fisheye datasets. 
Additionally, we present qualitative results of monocular depth estimation in Fig. \ref{fig:depth:qualy} and a 3D reconstruction on Fig. \ref{fig:depth:pc}.

\subsection{Semantic Segmentation}

\begin{table}[t]
\centering
\footnotesize	
	\begin{tabular}{c|cc|cc}
 	 & \multicolumn{1}{c|}{Dataset} & \multicolumn{1}{c|}{Kernel} & 
            mIoU     & mAcc\\ \hline
		\multicolumn{1}{c|}{\multirow{5}{*}{BL}}  & 
            \multicolumn{1}{c|}{\textit{Pers}} & Standard	& 
            33.32		& 42.35	\\ \cline{2-5}
            & \multicolumn{1}{c|}{\multirow{2}{*}{\textit{F195}}} & Standard   & 
            15.05		& 24.41	\\
		\multicolumn{1}{c|}{} & \multicolumn{1}{c|}{}         & Calibrated & 
            13.73		& 22.55	\\ \cline{2-5}
		  & \multicolumn{1}{c|}{\multirow{3}{*}{\textit{F165}}} & Standard	 & 
            15.12		& 24.36	\\
		\multicolumn{1}{c|}{} & \multicolumn{1}{c|}{}         & Calibrated & 
            13.23		& 21.46	\\
            \multicolumn{1}{c|}{} & \multicolumn{1}{c|}{}         & Rectified  &
            11.81		& 19.82	\\\hline \hline
		\multicolumn{1}{c|}{\multirow{5}{*}{FT}}  & 
            \multicolumn{1}{c|}{\multirow{2}{*}{\textit{F195}}}   & Standard   & 
            29.48		& 43.40	\\
		\multicolumn{1}{c|}{} & \multicolumn{1}{c|}{}         & Calibrated & 
            \textbf{30.90}		& \textbf{46.90}	\\ \cline{2-5}
		  & \multicolumn{1}{c|}{\multirow{3}{*}{\textit{F165}}} & Standard	 & 
            27.70		& 36.51	\\
		\multicolumn{1}{c|}{} & \multicolumn{1}{c|}{}         & Calibrated & 
            \textbf{27.89}		& \textbf{36.71}	\\
            \multicolumn{1}{c|}{} & \multicolumn{1}{c|}{}         & Rectified  & 
            21.99		& 29.61	\\ \hline
\end{tabular}
\caption{Semantic segmentation with standard and calibrated convolutions for U-Net neural network. BL: Base Line; FT: Fine Tuned. Best metric for each fisheye calibration is in bold.}
\label{tab:semSeg:UNET}
\end{table}

The second task that we evaluate with U-Net \cite{ronneberger2015u} is semantic segmentation. We use the same set-up and dataset than in the previous experiment. We train the network for 50 epochs, taking 75 hours to train, and then fine tune it 20 more epochs, taking between 5-7 more hours. 

\begin{figure}[t]
    \centering
    \subfloat[F165 mIoU]{\includegraphics[trim={0.9cm 0.9cm 0.9cm 0.9cm},width=0.24\textwidth]{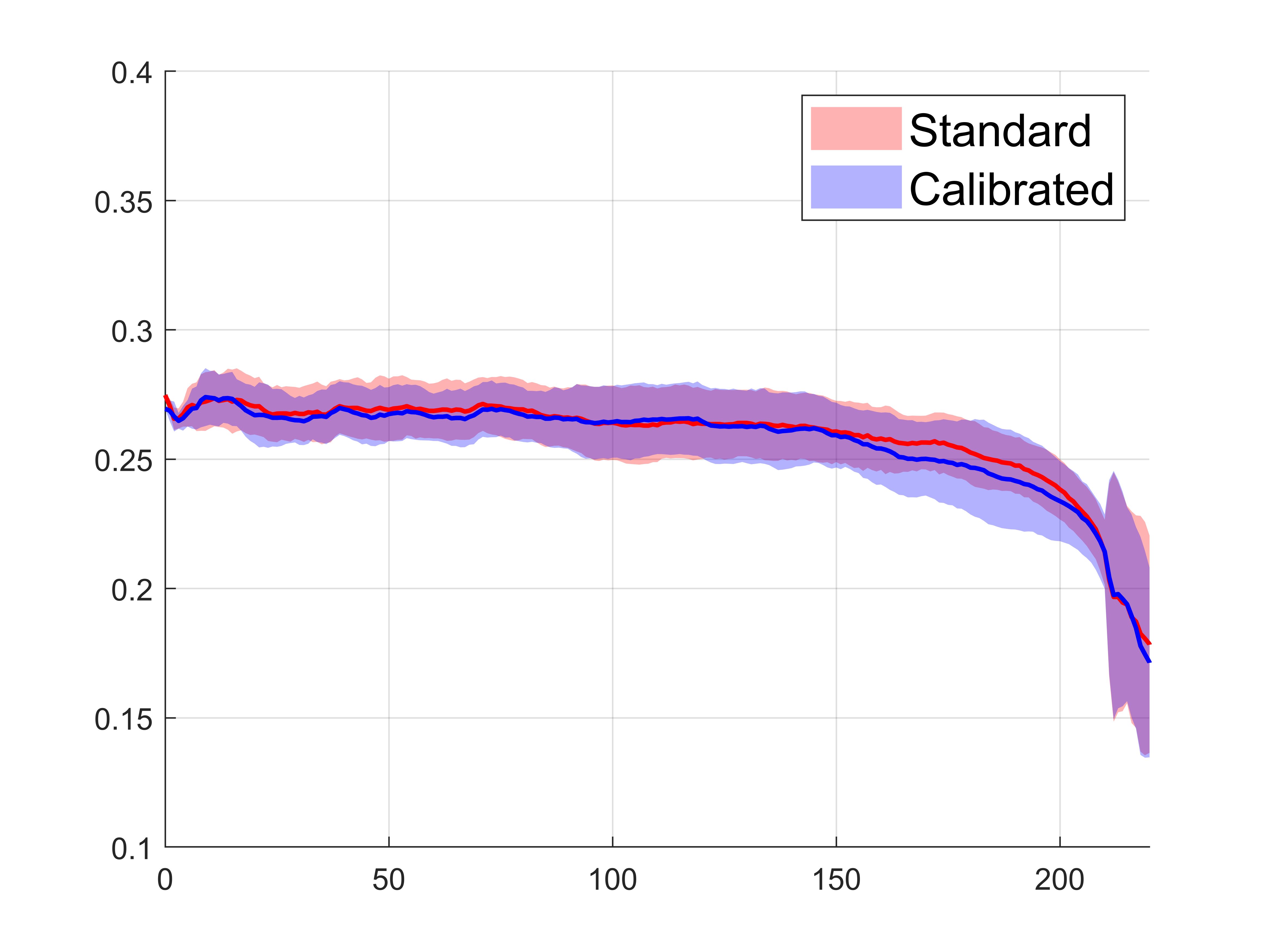}}
    \subfloat[F165 mAcc]{\includegraphics[trim={0.9cm 0.9cm 0.9cm 0.9cm},width=0.24\textwidth]{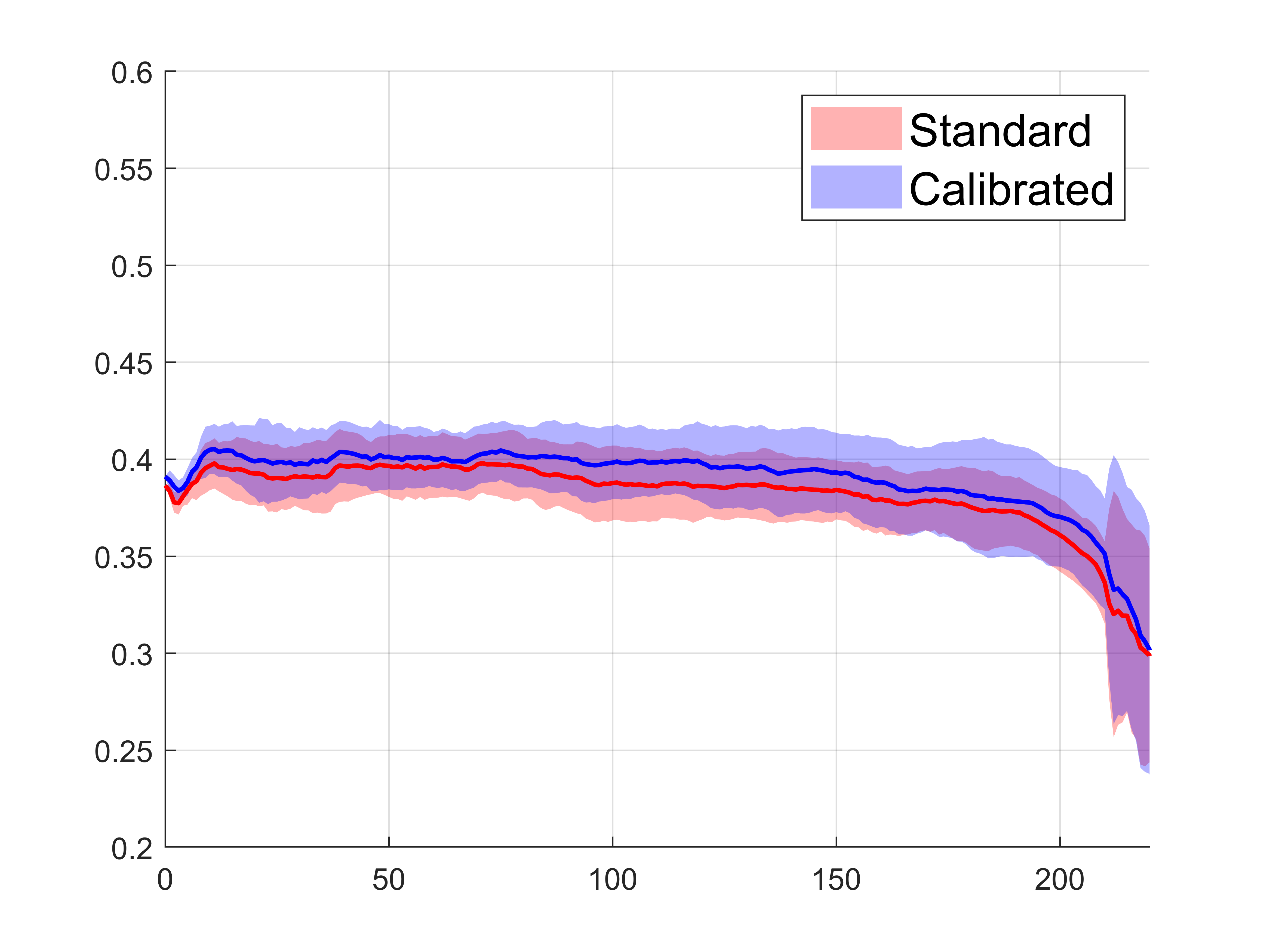}}
    \subfloat[F195 mIoU]{\includegraphics[trim={0.9cm 0.9cm 0.9cm 0.9cm},width=0.24\textwidth]{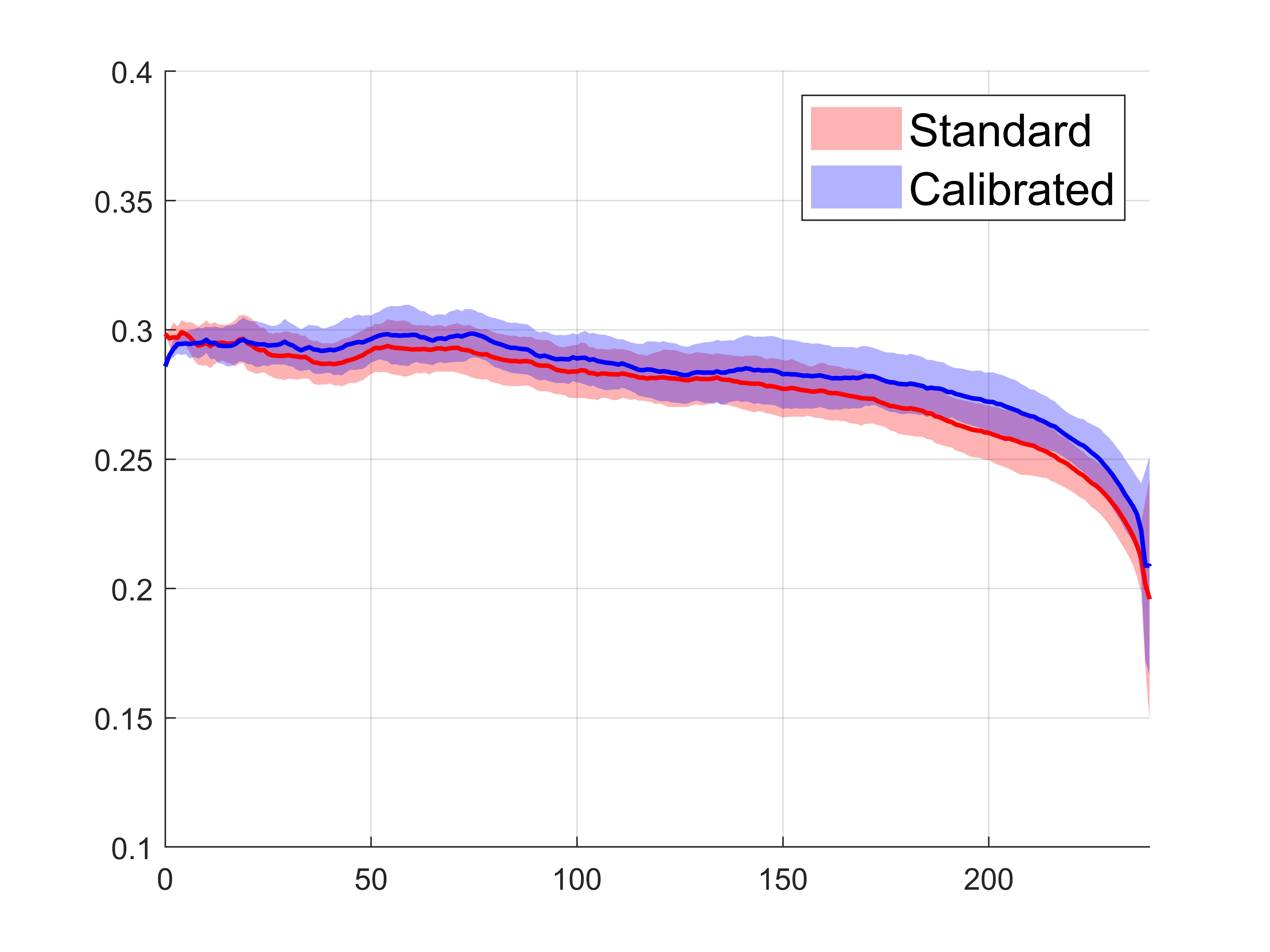}}
    \subfloat[F195 mAcc]{\includegraphics[trim={0.9cm 0.9cm 0.9cm 0.9cm},width=0.24\textwidth]{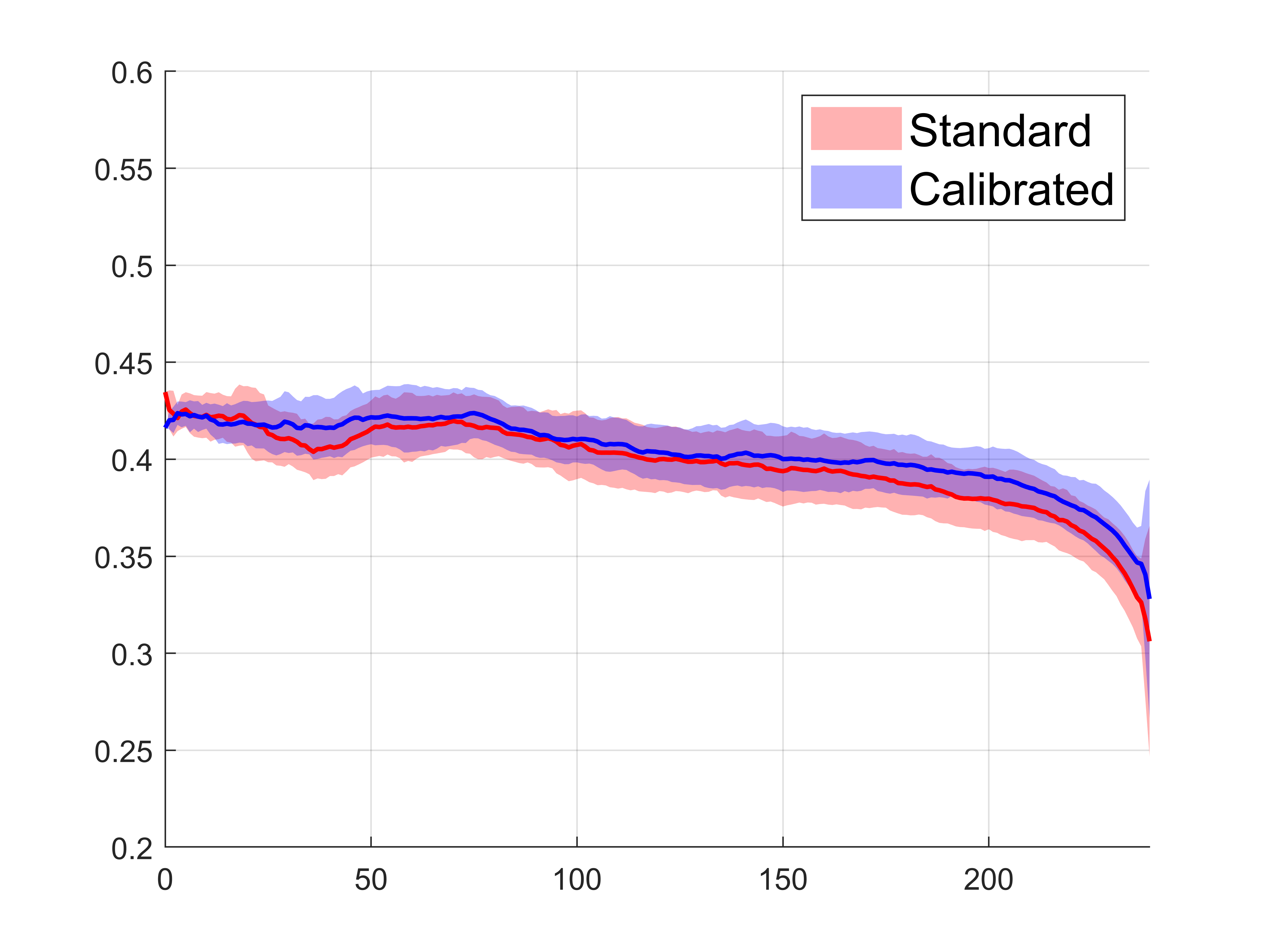}}
    \caption{Comparison and results of semantic segmentation with the U-Net like network with standard (red) and calibrated (blue) convolutions. The x-axis defines the distance of the pixels to the optical center and the y-axis the computed error, defined as mean and one standard deviation.}
    \label{fig:semantic:UNET}
\end{figure}

\begin{figure}[h]
    \centering
    \subfloat[F165 RGB]{\includegraphics[width=0.23\textwidth]{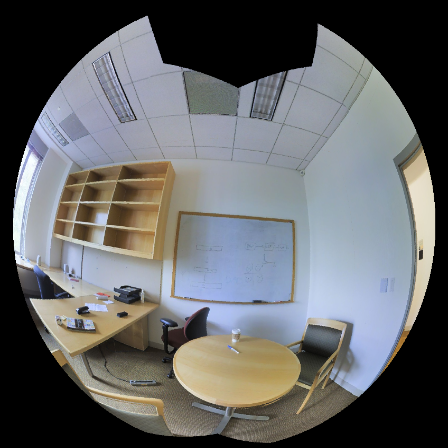}}
    \subfloat[F165 Standard]{\includegraphics[width=0.23\textwidth]{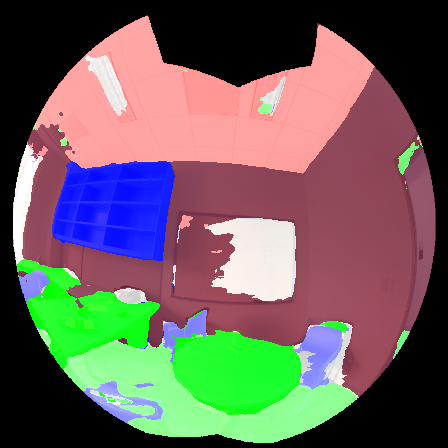}}
    \subfloat[F165 Calibrated]{\includegraphics[width=0.23\textwidth]{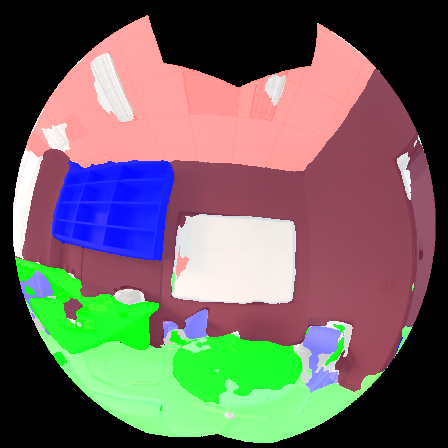}}
    \subfloat[F165 GT]{\includegraphics[width=0.23\textwidth]{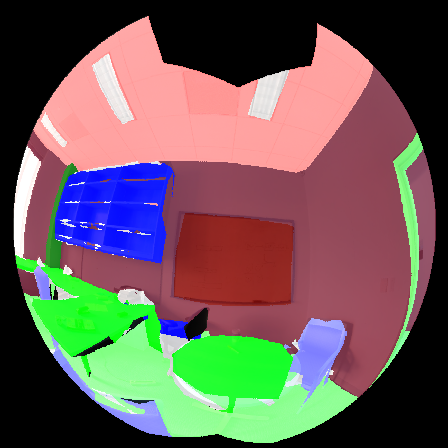}}\\

    \subfloat[F195 RGB]{\includegraphics[width=0.24\textwidth]{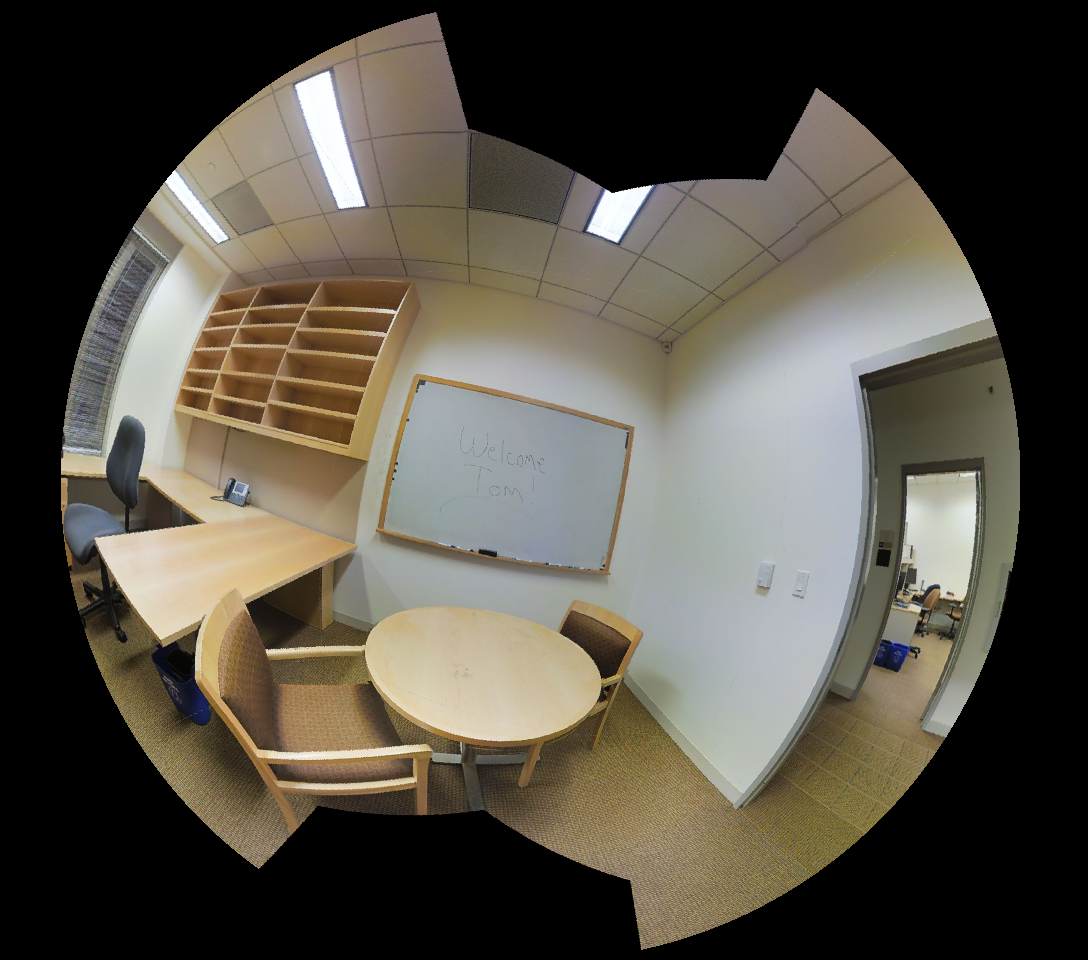}}
    \subfloat[F195 Standard]{\includegraphics[width=0.24\textwidth]{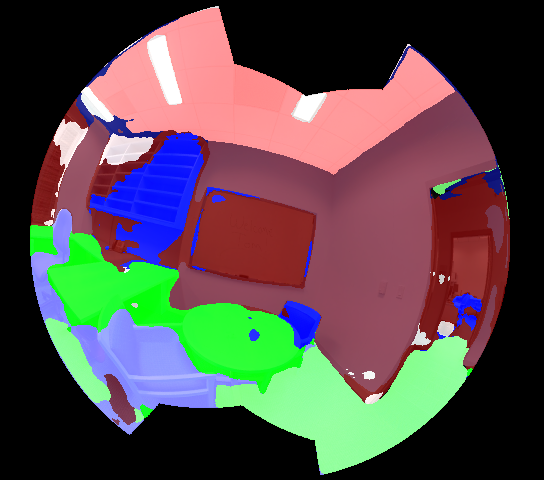}}
    \subfloat[F195 Calibrated]{\includegraphics[width=0.24\textwidth]{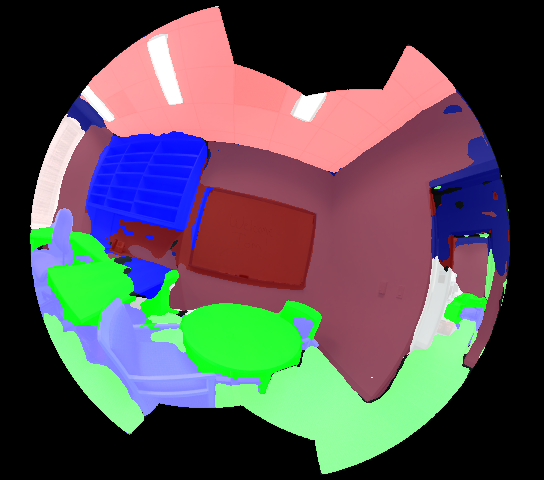}}
    \subfloat[F195 GT]{\includegraphics[width=0.24\textwidth]{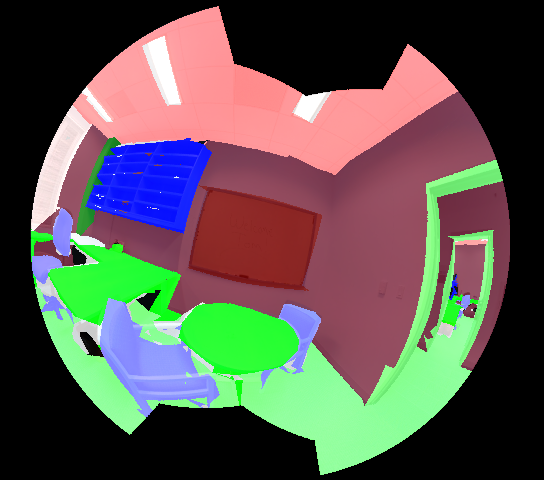}}\\
    \caption{Qualitative results of semantic segmentation on different fisheye calibrations. Each color represent a different class from the dataset.}
    \label{fig:semantic:qualy}
\end{figure}

Results of this experiment are shown in Tab. \ref{tab:semSeg:UNET}. The metrics used are the mean Intersection over Union (mIoU) and the mean Accuracy (mAcc) over all the classes, except the unknown class.
We also compute the metrics with respect the distance of each pixel to the principal point of the camera to evaluate the behaviour of the network with the increasing distortion of the image. These results are presented in Fig. \ref{fig:semantic:UNET} and qualitative results of semantic segmentation are presented in Fig. \ref{fig:semantic:qualy}.

\section{Discussion}

The experiments and results presented show that the transfer learning problem is still an open topic and difficult to achieve in the conversion from perspective to omnidirectional images. Both monocular depth estimation and semantic segmentation results present a great decrease of performance with the baseline weights of the network, being in some cases more accentuated when we use calibrated convolutions. However, after a short fine tune of the network, these results improve significantly, particularly with calibrated convolutions. 

From the results of depth estimation, we observe that the fine-tuned networks have slightly worse performance than the baseline with perspective images. The main difference is that with the fisheye images we cover a wider field of view, obtaining more information of the scene with the same number of images. 
In the comparison of convolutions, we observe that with wider fields of view, the calibrated convolutional kernels provide better performance, while with the smaller field of view, the performance is quite similar. However, when we make a deeper analysis of these results, in the Figure \ref{fig:depth:UNET} we observe the error distribution of the standard and calibrated convolutions. These results show that the estimation of the calibrated convolutions is more precise, with less dispersed error than the results of standard convolutions. Even if they are also affected by the increasing distortion of the fisheye images, the prediction is closer to the average error.

Regarding semantic segmentation experiments, the quantitative results from Tab. \ref{tab:semSeg:UNET} show that the performance with fisheye images decreases significantly from the perspective image case. This is to be expected, since the segmentation problem difficulty increases with the field of view, including more objects to segment in the same image, and distortion, changing the appearance of the same object in different locations in the image. However we mitigate the second problem with the calibrated convolutions, obtaining better results with our proposal than with standard convolutions consistently along the radius in most metrics, particularly in larger radius (see Fig. \ref{fig:semantic:UNET}). 
In addition, the qualitative results from Fig. \ref{fig:semantic:qualy} show significant differences in the performance between standard and calibrated convolutions. We can observe how the boundaries of objects and some details are better obtained with the calibrated convolutions than with the standard ones. 

These results and conclusions led us to believe that calibrated convolutions provide a faster domain adaptation of CNNs, that means, in the same training conditions with limited data, the calibrated convolutions provide better performance that the standard ones. The adaptation of networks trained on perspective images can be done with small datasets of fisheye images, achieving similar performance. 

\section{Conclusion}

In this article we have presented a novel implementation of deformable convolutional kernels taking into account the intrinsic calibration of fisheye cameras. Integrating the Kannala-Brandt projection model for revolution symmetry cameras in the kernel of convolutional neural networks, we obtain a domain adaptation mechanism to take advantage of previous works on perspective images and adapt these networks to work with fisheye cameras. 
On a similar approach, this work could also be extended to other projection models that take into account the calibration of omnidirectional cameras, such as the Scaramuzza's model \cite{scaramuzza2006}.

Results of the performed experiments show that the calibrated convolutions perform better than standard convolutions for domain adaptation. 
Besides, the impossibility of rectifying omnidirectional images of more than 180 degrees of field of view and the poor results obtained on the rectified ones increases the interest in studying how to adapt current deep learning methods to omnidirectional devices as the fisheye cameras.

A comparison with other methods, as CAM-Convs \cite{facil2019cam}, is not trivial. These works should be extended to other projection models, since currently only work on the pin-hole camera model. With a naive approach, including directly the Kannala-Brandt model to the CAM-Convs proposal, we observe abrupt changes in the feature maps, which make us believe that further research is needed. This new approach, the extension of methods as CAM-Convs to omnidirectional images, remains as future work.

\section*{ACKNOWLEDGMENT}

This work was supported by projects PID2021-125209OB-I00 and TED2021-129410B-I00 (MCIN/AEI/10.13039/501100011033 and FEDER/UE and NextGenerationEU/PRTR), and DGA 2022-2026 grant.

\bibliographystyle{unsrtnat}

\bibliography{references}

\begin{thebibliography}{33}
\providecommand{\natexlab}[1]{#1}
\providecommand{\url}[1]{\texttt{#1}}
\expandafter\ifx\csname urlstyle\endcsname\relax
  \providecommand{\doi}[1]{doi: #1}\else
  \providecommand{\doi}{doi: \begingroup \urlstyle{rm}\Url}\fi

\bibitem[Gao et~al.(2022)Gao, Yang, Shi, Wang, and Bai]{gao2022review}
Shaohua Gao, Kailun Yang, Hao Shi, Kaiwei Wang, and Jian Bai.
\newblock Review on panoramic imaging and its applications in scene
  understanding.
\newblock \emph{Transactions on Instrumentation and Measurement}, 71:\penalty0
  1--34, 2022.

\bibitem[Kumar et~al.(2018)Kumar, Milz, Witt, Simon, Amende, Petzold, Yogamani,
  and Pech]{kumar2018monocular}
Varun~Ravi Kumar, Stefan Milz, Christian Witt, Martin Simon, Karl Amende,
  Johannes Petzold, Senthil Yogamani, and Timo Pech.
\newblock Monocular fisheye camera depth estimation using sparse lidar
  supervision.
\newblock In \emph{International Conference on Intelligent Transportation
  Systems}, pages 2853--2858. IEEE, 2018.

\bibitem[Berenguel-Baeta et~al.(2023)Berenguel-Baeta, Bermudez-Cameo, and
  Guerrero]{berenguel2022fredsnet}
Bruno Berenguel-Baeta, Jesus Bermudez-Cameo, and Jose~J. Guerrero.
\newblock Fredsnet: Joint monocular depth and semantic segmentation with fast
  fourier convolutions from single panoramas.
\newblock In \emph{2023 IEEE International Conference on Robotics and
  Automation (ICRA)}, pages 6080--6086, 2023.
\newblock \doi{10.1109/ICRA48891.2023.10161142}.

\bibitem[Li et~al.(2022)Li, Guo, Yan, Huang, Duan, and Ren]{li2022omnifusion}
Yuyan Li, Yuliang Guo, Zhixin Yan, Xinyu Huang, Ye~Duan, and Liu Ren.
\newblock Omnifusion: 360 monocular depth estimation via geometry-aware fusion.
\newblock In \emph{Proceedings of Conference on Computer Vision and Pattern
  Recognition}, pages 2801--2810. IEEE/CVF, 2022.

\bibitem[Deng et~al.(2017)Deng, Yang, Qian, Wang, and Wang]{deng2017cnn}
Liuyuan Deng, Ming Yang, Yeqiang Qian, Chunxiang Wang, and Bing Wang.
\newblock Cnn based semantic segmentation for urban traffic scenes using
  fisheye camera.
\newblock In \emph{Intelligent Vehicles Symposium}, pages 231--236. IEEE, 2017.

\bibitem[Guerrero-Viu et~al.(2020)Guerrero-Viu, Fernandez-Labrador, Demonceaux,
  and Guerrero]{guerrero2020s}
Julia Guerrero-Viu, Clara Fernandez-Labrador, Cedric Demonceaux, and Jose~J
  Guerrero.
\newblock What’s in my room? object recognition on indoor panoramic images.
\newblock In \emph{International Conference on Robotics and Automation}, pages
  567--573. IEEE, 2020.

\bibitem[Haggui et~al.(2021)Haggui, Bayd, Magnier, and
  Aberkane]{haggui2021human}
Olfa Haggui, Hamza Bayd, Baptiste Magnier, and Arezki Aberkane.
\newblock Human detection in moving fisheye camera using an improved yolov3
  framework.
\newblock In \emph{International Workshop on Multimedia Signal Processing},
  pages 1--6. IEEE, 2021.

\bibitem[Toromanoff et~al.(2018)Toromanoff, Wirbel, Wilhelm, Vejarano,
  Perrotton, and Moutarde]{toromanoff2018end}
Marin Toromanoff, Emilie Wirbel, Fr{\'e}d{\'e}ric Wilhelm, Camilo Vejarano,
  Xavier Perrotton, and Fabien Moutarde.
\newblock End to end vehicle lateral control using a single fisheye camera.
\newblock In \emph{International Conference on Intelligent Robots and Systems},
  pages 3613--3619. IEEE/RSJ, 2018.

\bibitem[Baek et~al.(2018)Baek, Davies, Yan, and Rajkumar]{baek2018real}
Iljoo Baek, Albert Davies, Geng Yan, and Ragunathan~Raj Rajkumar.
\newblock Real-time detection, tracking, and classification of moving and
  stationary objects using multiple fisheye images.
\newblock In \emph{Intelligent vehicles symposium}, pages 447--452. IEEE, 2018.

\bibitem[Ni et~al.(2019)Ni, Ji, and Song]{ni2019vanishing}
Dejing Ni, Peng Ji, and Aiguo Song.
\newblock Vanishing point detection in corridor for autonomous mobile robots
  using monocular low-resolution fisheye vision.
\newblock \emph{Advances in Mechanical Engineering}, 11\penalty0 (10):\penalty0
  1687814019884767, 2019.

\bibitem[Perez-Yus et~al.(2019)Perez-Yus, Lopez-Nicolas, and
  Guerrero]{perez2019scaled}
Alejandro Perez-Yus, Gonzalo Lopez-Nicolas, and Jose~J Guerrero.
\newblock Scaled layout recovery with wide field of view rgb-d.
\newblock \emph{Image and Vision Computing}, 87:\penalty0 76--96, 2019.

\bibitem[Cohen et~al.(2018)Cohen, Geiger, K{\"o}hler, and
  Welling]{cohen2018spherical}
Taco~S Cohen, Mario Geiger, Jonas K{\"o}hler, and Max Welling.
\newblock Spherical cnns.
\newblock In \emph{International Conference on Learning Representations}, 2018.

\bibitem[Jiang et~al.(2019)Jiang, Huang, Kashinath, Marcus, Niessner,
  et~al.]{jiang2019spherical}
Chiyu Jiang, Jingwei Huang, Karthik Kashinath, Philip Marcus, Matthias
  Niessner, et~al.
\newblock Spherical cnns on unstructured grids.
\newblock \emph{arXiv preprint arXiv:1901.02039}, 2019.

\bibitem[Su and Grauman(2017)]{su2017learning}
Yu-Chuan Su and Kristen Grauman.
\newblock Learning spherical convolution for fast features from 360 imagery.
\newblock \emph{Advances in Neural Information Processing Systems}, 30, 2017.

\bibitem[Jiang et~al.(2021)Jiang, Sheng, Zhu, Dong, and
  Huang]{jiang2021unifuse}
Hualie Jiang, Zhe Sheng, Siyu Zhu, Zilong Dong, and Rui Huang.
\newblock Unifuse: Unidirectional fusion for 360 panorama depth estimation.
\newblock \emph{Robotics and Automation Letters}, 6\penalty0 (2):\penalty0
  1519--1526, 2021.

\bibitem[Zioulis et~al.(2019)Zioulis, Karakottas, Zarpalas, Alvarez, and
  Daras]{zioulis2019spherical}
Nikolaos Zioulis, Antonis Karakottas, Dimitrios Zarpalas, Federico Alvarez, and
  Petros Daras.
\newblock Spherical view synthesis for self-supervised 360 depth estimation.
\newblock In \emph{International Conference on 3D Vision}, pages 690--699.
  IEEE, 2019.

\bibitem[Jeon and Kim(2017)]{jeon2017active}
Yunho Jeon and Junmo Kim.
\newblock Active convolution: Learning the shape of convolution for image
  classification.
\newblock In \emph{Proceedings of Conference on Computer Vision and Pattern
  Recognition}, pages 4201--4209. IEEE/CVF, 2017.

\bibitem[Dai et~al.(2017)Dai, Qi, Xiong, Li, Zhang, Hu, and
  Wei]{dai2017deformable}
Jifeng Dai, Haozhi Qi, Yuwen Xiong, Yi~Li, Guodong Zhang, Han Hu, and Yichen
  Wei.
\newblock Deformable convolutional networks.
\newblock In \emph{Proceedings of the International Conference on Computer
  Vision}, pages 764--773. IEEE, 2017.

\bibitem[Zhuang et~al.(2022)Zhuang, Lu, Wang, Xiao, and Wang]{zhuang2022acdnet}
Chuanqing Zhuang, Zhengda Lu, Yiqun Wang, Jun Xiao, and Ying Wang.
\newblock Acdnet: Adaptively combined dilated convolution for monocular
  panorama depth estimation.
\newblock In \emph{Proceedings of the AAAI Conference on Artificial
  Intelligence}, volume~36, pages 3653--3661, 2022.

\bibitem[Fernandez-Labrador et~al.(2020)Fernandez-Labrador, Facil, Perez-Yus,
  Demonceaux, Civera, and Guerrero]{fernandez2020corners}
Clara Fernandez-Labrador, Jose~M Facil, Alejandro Perez-Yus, Cedric Demonceaux,
  Javier Civera, and Jose~J Guerrero.
\newblock Corners for layout: End-to-end layout recovery from 360 images.
\newblock \emph{Robotics and Automation Letters}, pages 1255--1262, 2020.

\bibitem[Tateno et~al.(2018)Tateno, Navab, and Tombari]{tateno2018distortion}
Keisuke Tateno, Nassir Navab, and Federico Tombari.
\newblock Distortion-aware convolutional filters for dense prediction in
  panoramic images.
\newblock In \emph{Proceedings of the European Conference on Computer Vision},
  pages 707--722. Springer, 2018.

\bibitem[Donahue et~al.(2014)Donahue, Jia, Vinyals, Hoffman, Zhang, Tzeng, and
  Darrell]{donahue2014decaf}
Jeff Donahue, Yangqing Jia, Oriol Vinyals, Judy Hoffman, Ning Zhang, Eric
  Tzeng, and Trevor Darrell.
\newblock Decaf: A deep convolutional activation feature for generic visual
  recognition.
\newblock In \emph{International conference on machine learning}, pages
  647--655. PMLR, 2014.

\bibitem[Sharif~Razavian et~al.(2014)Sharif~Razavian, Azizpour, Sullivan, and
  Carlsson]{sharif2014cnn}
Ali Sharif~Razavian, Hossein Azizpour, Josephine Sullivan, and Stefan Carlsson.
\newblock Cnn features off-the-shelf: an astounding baseline for recognition.
\newblock In \emph{Proceedings of Conference on Computer Vision and Pattern
  Recognition workshops}, pages 806--813. IEEE/CVF, 2014.

\bibitem[Artizzu et~al.(2023)Artizzu, Allibert, and
  Demonceaux]{artizzu2023omni}
Charles-Olivier Artizzu, Guillaume Allibert, and C{\'e}dric Demonceaux.
\newblock Omni-conv: Generalization of the omnidirectional distortion-aware
  convolutions.
\newblock \emph{Journal of Imaging}, 9\penalty0 (2):\penalty0 29, 2023.

\bibitem[Meng et~al.(2021)Meng, Xiao, Zhou, Li, and Zhou]{meng2021distortion}
Ming Meng, Likai Xiao, Yi~Zhou, Zhaoxin Li, and Zhong Zhou.
\newblock Distortion-aware room layout estimation from a single fisheye image.
\newblock In \emph{International Symposium on Mixed and Augmented Reality},
  pages 441--449. IEEE, 2021.

\bibitem[Facil et~al.(2019)Facil, Ummenhofer, Zhou, Montesano, Brox, and
  Civera]{facil2019cam}
Jose~M Facil, Benjamin Ummenhofer, Huizhong Zhou, Luis Montesano, Thomas Brox,
  and Javier Civera.
\newblock Cam-convs: Camera-aware multi-scale convolutions for single-view
  depth.
\newblock In \emph{Proceedings of the Conference on Computer Vision and Pattern
  Recognition}, pages 11826--11835. IEEE/CVF, 2019.

\bibitem[Kannala and Brandt(2006)]{kannala2006generic}
Juho Kannala and Sami~S Brandt.
\newblock A generic camera model and calibration method for conventional,
  wide-angle, and fish-eye lenses.
\newblock \emph{Transactions on Pattern Analysis and Machine Intelligence},
  pages 1335--1340, 2006.

\bibitem[Scaramuzza et~al.(2006)Scaramuzza, Martinelli, and
  Siegwart]{scaramuzza2006}
Davide Scaramuzza, Agostino Martinelli, and Roland Siegwart.
\newblock A flexible technique for accurate omnidirectional camera calibration
  and structure from motion.
\newblock In \emph{International Conference on Computer Vision Systems}, pages
  45--45. IEEE, 2006.

\bibitem[Ronneberger et~al.(2015)Ronneberger, Fischer, and
  Brox]{ronneberger2015u}
Olaf Ronneberger, Philipp Fischer, and Thomas Brox.
\newblock U-net: Convolutional networks for biomedical image segmentation.
\newblock In \emph{Medical Image Computing and Computer-Assisted Intervention},
  pages 234--241. Springer, 2015.

\bibitem[Hochreiter and Schmidhuber(1997)]{hochreiter1997long}
Sepp Hochreiter and J{\"u}rgen Schmidhuber.
\newblock Long short-term memory.
\newblock \emph{Neural computation}, 9\penalty0 (8):\penalty0 1735--1780, 1997.

\bibitem[Vaswani et~al.(2017)Vaswani, Shazeer, Parmar, Uszkoreit, Jones, Gomez,
  Kaiser, and Polosukhin]{vaswani2017attention}
Ashish Vaswani, Noam Shazeer, Niki Parmar, Jakob Uszkoreit, Llion Jones,
  Aidan~N Gomez, {\L}ukasz Kaiser, and Illia Polosukhin.
\newblock Attention is all you need.
\newblock In \emph{Conference on Neural Information Processing Systems}, 2017.

\bibitem[Armeni et~al.(2017)Armeni, Sax, Zamir, and Savarese]{armeni2017joint}
Iro Armeni, Sasha Sax, Amir~R Zamir, and Silvio Savarese.
\newblock Joint 2d-3d-semantic data for indoor scene understanding.
\newblock \emph{arXiv preprint arXiv:1702.01105}, 2017.

\bibitem[Zioulis et~al.(2018)Zioulis, Karakottas, Zarpalas, and
  Daras]{zioulis2018omnidepth}
Nikolaos Zioulis, Antonis Karakottas, Dimitrios Zarpalas, and Petros Daras.
\newblock Omnidepth: Dense depth estimation for indoors spherical panoramas.
\newblock In \emph{European Conference on Computer Vision}, pages 448--465.
  Springer, 2018.

\end{thebibliography}

\end{document}